\documentclass[journal]{IEEEtran}

\ifCLASSINFOpdf
\else
\fi

\usepackage{setspace}
\usepackage{duckuments} 
\usepackage{times}
\usepackage{subfig}
\usepackage{graphicx}
\usepackage{amsmath,amsthm}
\usepackage{amssymb,wasysym}
\usepackage{mathrsfs}
\usepackage{cite}
\usepackage{xcolor}
\usepackage{url}
\usepackage[lined,linesnumbered,ruled]{algorithm2e}
\usepackage{algorithmic}
\usepackage{wrapfig}
\usepackage{verbatim}
\usepackage{dsfont}
\usepackage{multirow}

\ifodd 0
\newcommand{\rev}[1]{{\color{green}#1}} 
\newcommand{\needrev}[1]{{\color{red}#1}} 
\newcommand{\copyrev}[1]{{\color{blue}#1}} 
\newcommand{\waitrev}[1]{{\color{orange}#1}} 
\newcommand{\peerrev}[1]{{\color{pink}#1}} 
 
\else
\newcommand{\rev}[1]{#1}
\newcommand{\needrev}[1]{#1}
\newcommand{\copyrev}[1]{#1}
\newcommand{\waitrev}[1]{#1}
\newcommand{\peerrev}[1]{#1}
\fi

\begin{document}

\newcommand{\name}{HFedMoE\xspace}

\title{\name: Resource-aware Heterogeneous Federated Learning with Mixture-of-Experts}

\author{Zihan Fang, Zheng Lin, Senkang Hu, Yanan Ma, Yihang Tao, Yiqin Deng,~\IEEEmembership{Member,~IEEE}, \\
Xianhao Chen,~\IEEEmembership{Member,~IEEE} and Yuguang Fang,~\IEEEmembership{Fellow,~IEEE}

\thanks{{The research work described in this paper was conducted in the JC STEM Lab of Smart City funded by The Hong Kong Jockey Club Charities Trust under Contract 2023-0108.  This work was also supported in part by the Hong Kong SAR Government under the Global STEM Professorship and Research Talent Hub, and in part by the Hong Kong Innovation and Technology Commission under InnoHK Project CIMDA. The work of Yiqin Deng was supported in part by the National Natural Science Foundation of China under Grant No. 62301300. The work of Xianhao Chen was supported in part by HKU-SCF FinTech Academy R\&D Funding.}
}

\thanks{Z. Fang, S. Hu, Y. Ma, Y. Tao, Y. Deng and Y. Fang are with Hong Kong JC STEM Lab of Smart City and Department of Computer Science, City University of Hong Kong, Kowloon, Hong Kong SAR, China (e-mail: zihanfang3-c@my.cityu.edu.hk; senkang.forest@my.cityu.edu.hk; yananma8-c@my.cityu.edu.hk; yihang.tommy@my.cityu.edu.hk; yiqideng@cityu.edu.hk; my.fang@cityu.edu.hk).}
\thanks{Z. Lin and X. Chen are with the Department of Electrical and Electronic Engineering, The University of Hong Kong, Pok Fu Lam, Hong Kong, China (e-mail: linzheng@eee.hku.hk; xchen@eee.hku.hk).}

}


\maketitle

\begin{abstract}
While federated learning (FL) enables fine-tuning of large language models (LLMs) without compromising data privacy, the substantial size of an LLM renders on-device training impractical for resource-constrained clients, such as mobile devices. Thus, Mixture-of-Experts (MoE) models have emerged as a computation-efficient solution, which activates only a sparse subset of experts during model training to reduce computing burden without sacrificing performance. 
Though integrating MoE into FL fine-tuning holds significant potential, it still encounters three key challenges: i) selecting appropriate experts for clients remains challenging due to the lack of a reliable metric to measure each expert’s impact on local fine-tuning performance, ii) the heterogeneous computing resources across clients severely hinder MoE-based LLM fine-tuning, as dynamic expert activations across diverse input samples can overwhelm resource-constrained devices, and iii) \needrev{client-specific expert subsets and routing preference undermine global aggregation, where misaligned expert updates and inconsistent gating networks introduce destructive interference.}
To address these challenges, we propose \name, a heterogeneous MoE-based FL fine-tuning framework that customizes a subset of experts to each client for computation-efficient LLM fine-tuning.  
Specifically, \name identifies the expert importance based on its contributions to fine-tuning performance, and then adaptively selects a subset of experts from an information bottleneck perspective to align with each client’s computing budget. A sparsity-aware model aggregation strategy is also designed to aggregate the actively fine-tuned experts \waitrev{and gating parameters with importance-weighted contributions}. 
Extensive experiments demonstrate that \name outperforms state-of-the-art benchmarks in training accuracy and convergence speed.
\end{abstract}


\IEEEpeerreviewmaketitle

\vspace{-0.2cm}
\section{Introduction} \label{sec:introduction}
Recently, large language models (LLMs) such as GPT~\cite{achiam2023gpt,lin2023pushing,duan2025leed}, LLaMA~\cite{touvron2023llama,lin2024splitlora}, and DeepSeek~\cite{liu2024deepseek} have attracted significant attention from both academia and industry due to their superior ability in handling high-complexity and large-scale datasets~\cite{fang2024automated,lin2025hsplitlora,duan2025llm,fang2025dynamic}. 
\peerrev{LLMs generally follow a two-stage training procedure. First, LLMs are pre-trained on massive text corpora (e.g., Wikipedia) to learn universal linguistic and semantic representations. Second, the pre-trained LLMs are adapted to a downstream task with task-specific data.}
However, the excessive data required for fine-tuning LLMs raises serious privacy concerns, \peerrev{posing a substantial barrier to the implementation of LLM fine-tuning}. For instance, clients are often reluctant to share their privacy-sensitive data, such as personal healthcare records or financial information~\cite{karargyris2023federated}, to train a shared LLM.

To address the above issue, federated learning (FL)~\cite{mcmahan17a, luo2023optimization,hu2024accelerating,zhang2024satfed,bonawitz2019towards,lin2024fedsn,zhang2025lcfed} has emerged as a viable alternative, \peerrev{which enables collaborative training across clients without exposing raw data}. 
The standard FL procedure for LLM fine-tuning comprises three phases: i) Each client independently fine-tunes its LLM using local private dataset; ii) All clients upload their locally updated LLM to a central server for model aggregation (e.g., via weighted averaging); and iii) the aggregated LLM is distributed back to clients before the next training round. 
Despite its privacy-preserving nature, fine-tuning LLMs via FL is computationally prohibitive on resource-constrained devices~\cite{wang2024flora,lin2024adaptsfl,jiang2023computation, lin2024efficient,lin2025hasfl}. 
Commercial-grade on-device GPUs (e.g., \needrev{NVIDIA Jetson Orin}) \peerrev{lack sufficient processing capacities} to fine-tune LLMs like LLaMA-2 7B, which demands nearly \needrev{60} GB of GPU memory\footnote{In this paper, we use GPU memory footprint as a proxy to quantify the computing resource consumption~\cite{dhar2024empirical, xu2025edgellm, qu2025mobile}. Since memory-intensive operations, such as \rev{intermediate state memory storage} and large matrix multiplication, dominate LLM fine-tuning, the GPU memory footprint can reflect the computing demands of LLM fine-tuning. This makes it an effective measure for fine-tuning feasibility on resource-constrained client devices.} and 187.9 TFLOPs per 4K tokens~\cite{dai2024deepseekmoe}, far exceeding the capabilities of mobile client devices. 
To reduce the LLM fine-tuning overhead, parameter-efficient fine-tuning (PEFT) techniques such as LoRA~\cite{hu2021lora} and Adapters~\cite{houlsby2019parameter} have been proposed. 
While PEFT significantly reduces the number of trainable parameters, \peerrev{it entails full forward and backward propagation through the entire LLM. Due to the large model size of LLMs, client devices still do not sufficiently meet the computing demands for PEFT LLM via FL.} Even with LoRA, fine-tuning LLaMA-2 7B requires over 20 GB of GPU memory, which easily overwhelms most mobile client devices~\cite{lin2025hierarchical}.


\begin{figure}[t!]
\centering
\includegraphics[width=8.5cm]{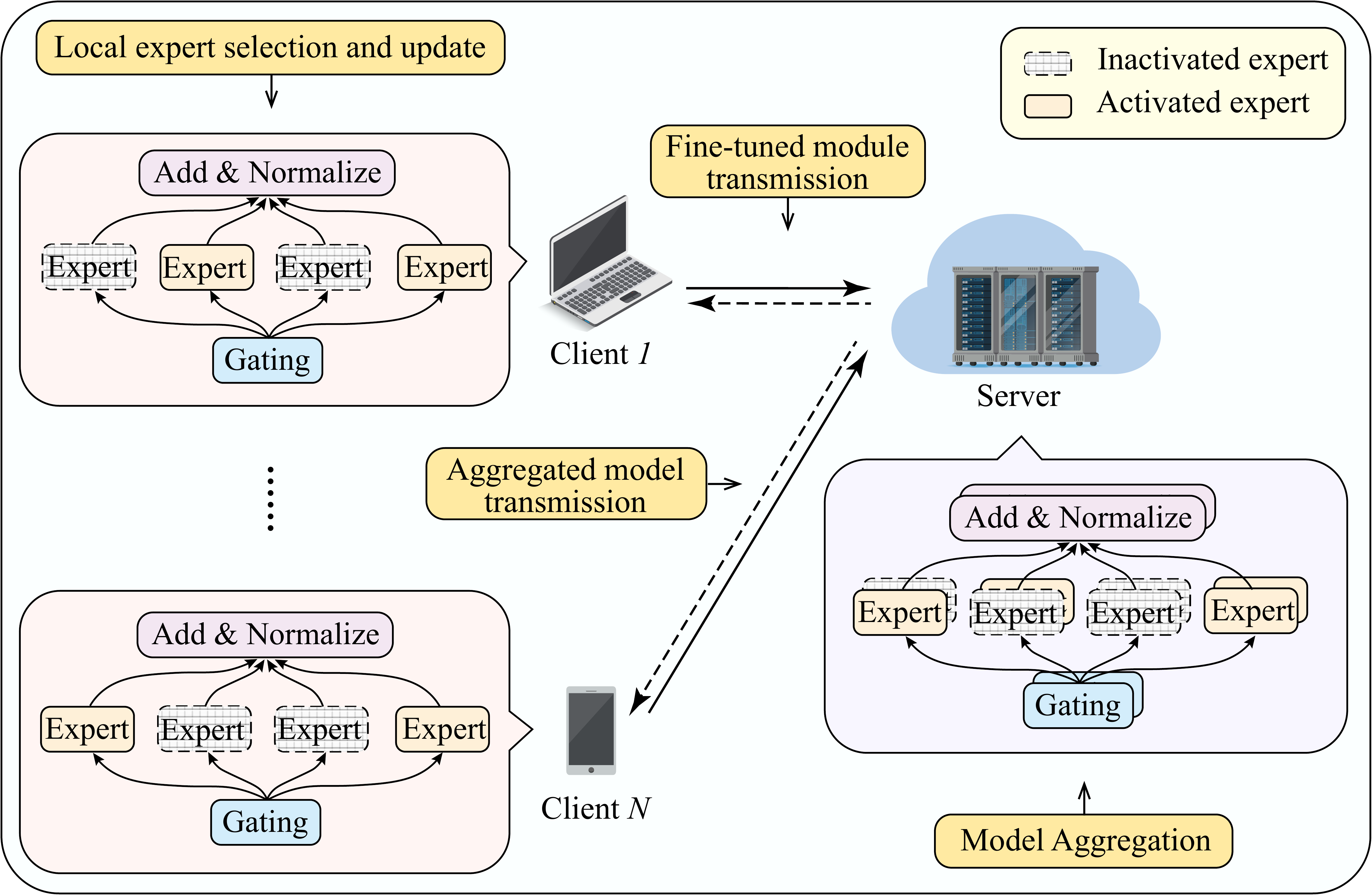}
\caption{The workflow for fine-tuning MoE-based LLM via FL across clients.}
\label{fig:fl_workflow}
\vspace{-2ex}
\end{figure}

Mixture-of-experts (MoE)~\cite{dai2024deepseekmoe, lu2024not, fedus2022switch} offers a structurally efficient solution to fine-tune LLMs on resource-limited mobile client devices via FL. 
Different from PEFT methods that require full-model forward and backward propagation, MoE activates only a small number of experts (e.g., top-1 or top-2) for each input, while keeping the rest inactive. 
This sparse activation \peerrev{preserves the full representational capacity of the model} while substantially reducing the computational cost~\cite{dai2024deepseekmoe, fedus2022switch, muzio2024seer}, particularly during gradient computation and weight updates, which dominate training cost in LLM fine-tuning~\cite{du2024sida}. 
\peerrev{Beyond computational benefits, MoE is naturally well-suited for FL, as input-dependent expert routing enables client-specific specialization, facilitating personalized model adaptation. 
}
Fig.~\ref{fig:fl_workflow} illustrates how MoE integrates with FL for LLM fine-tuning. Each client device fine-tunes a personalized subset of experts selected by a gating network. \peerrev{Then, all client devices upload the gating network and experts to the server for model aggregation to update the LLM.}
This design can achieve comparable performance of large dense models (e.g., LLaMA-2 7B) while consuming less than 40\% of the computation~\cite{dai2024deepseekmoe}.

While integrating MoE into FL holds significant promise, it encounters several critical challenges. 
First, selecting appropriate experts for each client is non-trivial, due to client heterogeneity in local datasets. 
The performance of individual experts varies significantly across clients~\cite{lu2024not, guo2021pfl}, \rev{i.e., certain experts may generalize well to specific client datasets, they may offer negligible benefit to others}. 
As a result, a shared gating network for expert selection fail to capture these client-specific differences, leading to suboptimal expert utilization. \peerrev{This misalignment not only wastes computing resources but also compromises the overall performance of the model.} 
Second, heterogeneous computing resources across clients severely limit the deployment of MoE-based fine-tuning via FL. 
\peerrev{The divergent expert activation across input tokens leads to an increased number of experts concurrently activated within one training batch~\cite{fedus2022switch, lewis2021base}, \rev{varying from a few for simple samples to many for complicated ones.}}
This can easily exceed the computing capabilities of resource-constrained client devices, thus resulting in fine-tuning failures and severely lowering the overall
training efficiency.
\waitrev{Third, varying computing resources and data distributions results in client-specific expert selections and routing preferences, making model aggregation in MoE-based FL particularly challenging.
Each client fine-tunes only a subset of experts aligned with its local data, leaving other experts undertrained and making clients’ gating networks to favor distinct expert subsets. Such misalignment causes severe interference during aggregation, thus degrading the generalization of the global model.}
We will empirically conduct measurement studies in Sec.~\ref{sec:motivation} to investigate these challenges.

To tackle the above challenges, we propose a \underline{h}eterogeneous \underline{fed}erated learning framework for \underline{MoE}-based LLM fine-tuning, named \name, to selectively fine-tune the subset of experts for each client under heterogeneous \rev{edge computing systems}. 
\peerrev{First, to select appropriate experts for each client device, we introduce an expert importance identification scheme that quantifies each expert's contribution to the fine-tuning performance of diverse clients.}
Second, to accommodate the heterogeneous computing budgets of client devices, we propose a resource-aware expert selection method that dynamically selects the subset of critical experts to align with each device’s computing budget \needrev{from the information bottleneck perspective}, enhancing training efficiency without compromising model performance. 
\waitrev{Finally, to mitigate aggregation discrepancies caused by \rev{partial expert updates and diverse routing preferences}, we design a sparsity-aware model aggregation strategy that updates only the experts actively trained on clients, while aggregating gating parameters with importance-weighted contributions, enhancing the generalization of the global MoE model with structural heterogeneity.}
The key contributions of this paper are summarized as follows.
\begin{itemize}
  \item We propose \name, a MoE-based FL fine-tuning framework to customize the subset of experts to each client for model training, enabling efficient LLM fine-tuning under heterogeneous edge computing capabilities.
  \item We design an expert importance identification scheme to quantify experts' contributions to local training performance, so as to prioritize critical  experts for fine-tuning.
  \item We devise resource-aware expert selection to dynamically select a subset of critical experts to align with each device’s computing budget during fine-tuning.
  \item \waitrev{We develop a sparsity-aware model aggregation strategy to explicitly handle partial expert updates and routing inconsistency, mitigating performance degradation caused by structural heterogeneity across clients.} 
  \item We empirically validate the fine-tuning performance of \name with extensive experiments, demonstrating the superiority of \name over state-of-the-art frameworks in both \peerrev{model accuracy and convergence speed}.
\end{itemize}

\copyrev{This paper is organized as follows.
Sec.~\ref{sec:motivation} motivates the design of \name by revealing the challenges of incorporating MoE in FL network.
Sec.~\ref{sec:design} elaborates on the framework design, followed by performance evaluation in Sec.~\ref{sec:simulation}.
Related works and technical limitations are discussed in Sec.~\ref{sec:related_work}.
Finally, conclusions are presented in Sec.~\ref{sec:conclusion}.
}

\vspace{-2ex}
\section{Challenges and Motivation}\label{sec:motivation}
In this section, we conduct extensive pilot measurement studies to elaborate the key challenges of fine-tuning MoE-based LLM in federated learning, which motivates the design of \name.

\vspace{-2ex}
\subsection{Suboptimal Expert Utilization} \label{sec:mtv_expert}
For MoE-based FL fine-tuning, expert routing is typically controlled by a shared global gating network~\cite{reisser2021federated, dun2023fedjets}. Although this enables centralized coordination, it fails to capture the heterogeneity of client-specific data distributions.
Since each expert focuses on distinct semantic features~\cite{reisser2021federated}, its contribution (i.e., the task-relevant information it provides) vary significantly among clients~\cite{lu2024not, guo2021pfl}.
However, the global gating network is optimized by average routing performance across all clients, ignoring client-specific preferences. This mismatch causes critical experts for some clients to be underutilized, while frequently activated experts may contribute little locally, ultimately leading to suboptimal expert utilization and degrading training performance.

\begin{figure}[t]
\centering
\subfloat[Client-specific expert selections \label{fig:mtv_expert_selection}]{
\includegraphics[width=0.49\linewidth]{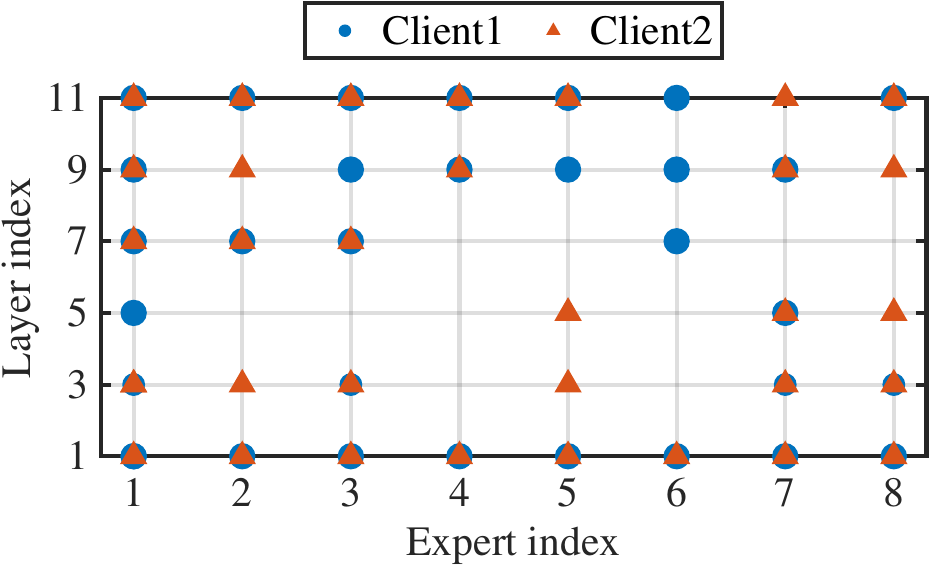}
}
\subfloat[Comparison after aggregation \label{fig:mtv_expert_performance}]{
\includegraphics[width=0.49\linewidth]{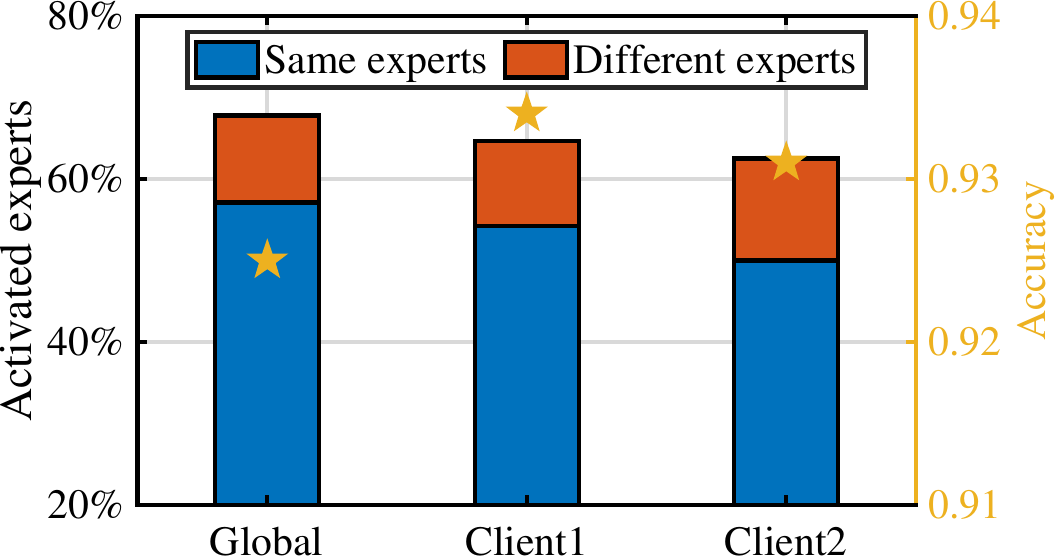}
}
\caption{The expert activations on AGNews dataset with global or local expert routing across clients, under a batch size of 4.
}
\label{fig:mtv_expert}
\vspace{-3ex}
\end{figure}

To investigate the impact of suboptimal expert utilization, we conduct a motivating study using Switch Transformer~\cite{fedus2022switch} with 8 experts per layer on AGNew dataset~\cite{zhang2015character}. We compare global routing (i.e., all clients share a single gating network) with client-specific routing (i.e., each client independently trains its gating network). Fig.~\ref{fig:mtv_expert_selection} shows that \needrev{local expert selections} diverge significantly across clients, indicating the sensitivity of experts to local dataset. Furthermore, Fig.\ref{fig:mtv_expert_performance} reveals that gating network aggregation changes local expert selections and fails to accurately prioritize vital experts to specific clients, thereby degrading local fine-tuning accuracy (yellow star in the figure) These observations underscore the necessity of identifying expert importance and selectively activating experts specific to each client for enhancing both efficiency and personalization.

\begin{figure}[t]
\centering
\subfloat[Activated expert proportion \label{fig:mtv_computing_activation}]{
\includegraphics[width=0.49\linewidth]{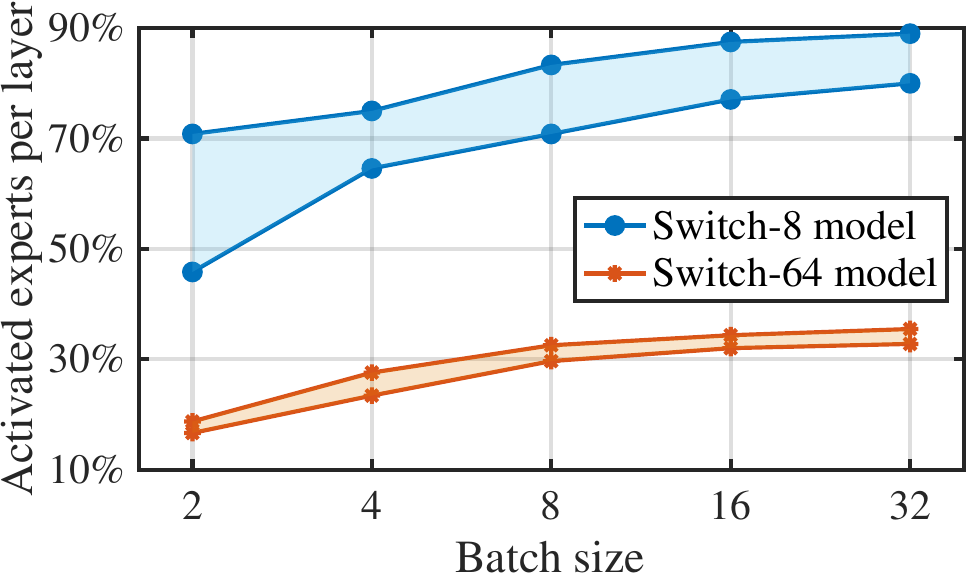}
}
\subfloat[Converged performance \label{fig:mtv_computing_hetero}]{
\includegraphics[width=0.49\linewidth]{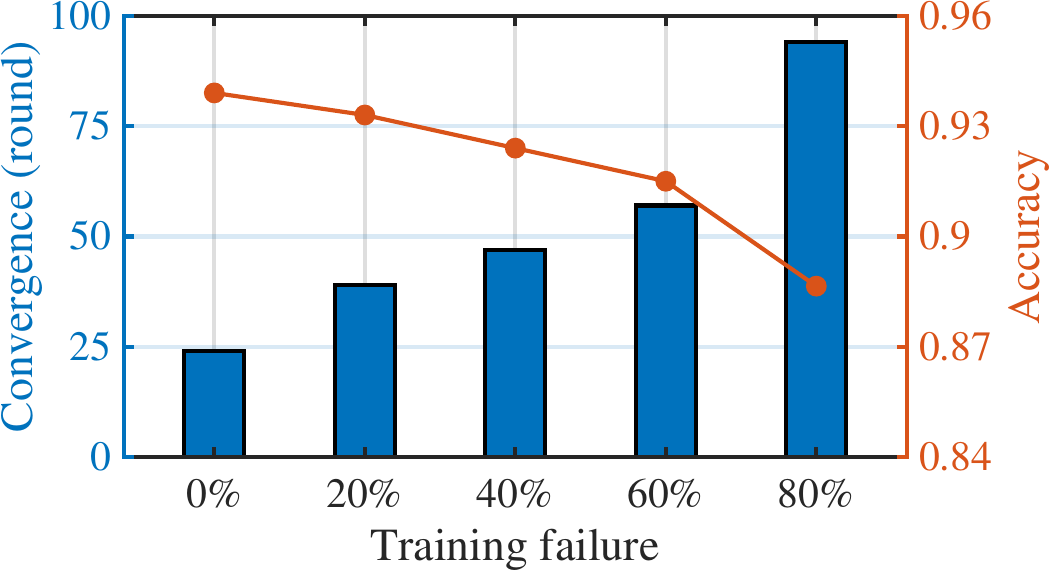}
}
\caption{
The activated expert proportion within varying batch sizes and the converged performance (training round and test accuracy) under varying training failure rates.
}
\label{fig:mtv_computing}
\vspace{-2ex}
\end{figure}


\vspace{-2ex}
\subsection{Heterogeneous Computing Resources} \label{sec:mtv_computing}

\rev{While MoE models enable sparse expert activation for efficient LLM fine-tuning}, the heterogeneity in computing capabilities across client devices, stemming from differences in hardware configurations and deployment environments~\cite{jhang2021challenges, zhang2023resource, lin2024fedsn, zhu2024esfl}, poses significant challenges. 
Despite the selective activation of top-1 expert per token, variations in expert selection from one token to another \rev{across samples within a training batch} often lead to varying number of experts being activated concurrently~\cite{fedus2022switch, lewis2021base}. 
This undermines the intended sparsity design of MoE and imposes a heavy computing burden on resource-constrained clients (e.g., those equipped with commercial-grade GPUs). 
\peerrev{Consequently, this mismatch in resource demand and availability makes on-device fine-tuning of FL infeasible.} For instance, fine-tuning the DeepSeekMoE-16B model requires 74.4T FLOPs per 4K tokens~\cite{dai2024deepseekmoe}, still exceeding the processing capabilities of most client GPUs and thus degrading the overall training efficiency.

To better understand the impact of heterogeneous client computing capabilities on FL training performance, we conduct motivating experiments using Switch Transformer with 8 or 64 experts per layer on the AGNews dataset. 
By analyzing expert activations \peerrev{at a fixed sequence length of 128} under varying batch sizes, Fig.~\ref{fig:mtv_computing_activation} reveals that even with top-1 routing and a small batch size of 2, over 15\% of the 64 experts are activated—substantially exceeding the intended selection of one expert per layer. 
Additionally, Fig.~\ref{fig:mtv_computing_hetero} shows that the clients with limited resources frequently fail to complete local LLM fine-tuning, and severe disparities in computational capacity among clients significantly degrade the overall convergence and stability of collaborative training. 
These results underscore the negative impact of \needrev{inefficient token-wise expert selection} for resource-constrained MoE fine-tuning, motivating the design of a resource-aware expert selection mechanism.

\begin{figure}[t]
\centering
\subfloat[The selected experts in 4 clients \label{fig:mtv_aggregation_overlap}]{
\includegraphics[width=0.49\linewidth]{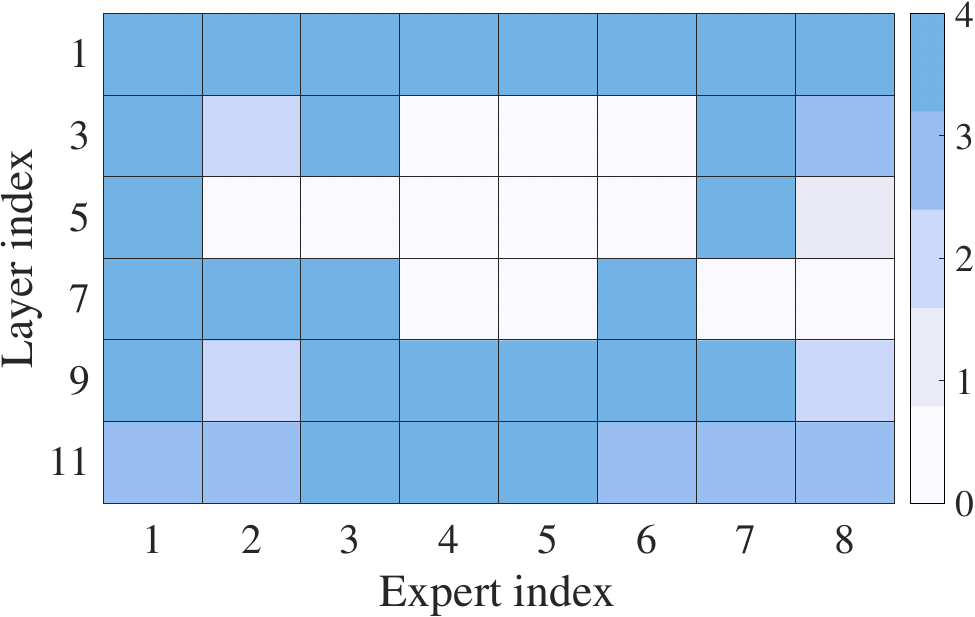}
}
\subfloat[Aggregation performance \label{fig:mtv_aggregation_performance}]{
\includegraphics[width=0.49\linewidth]{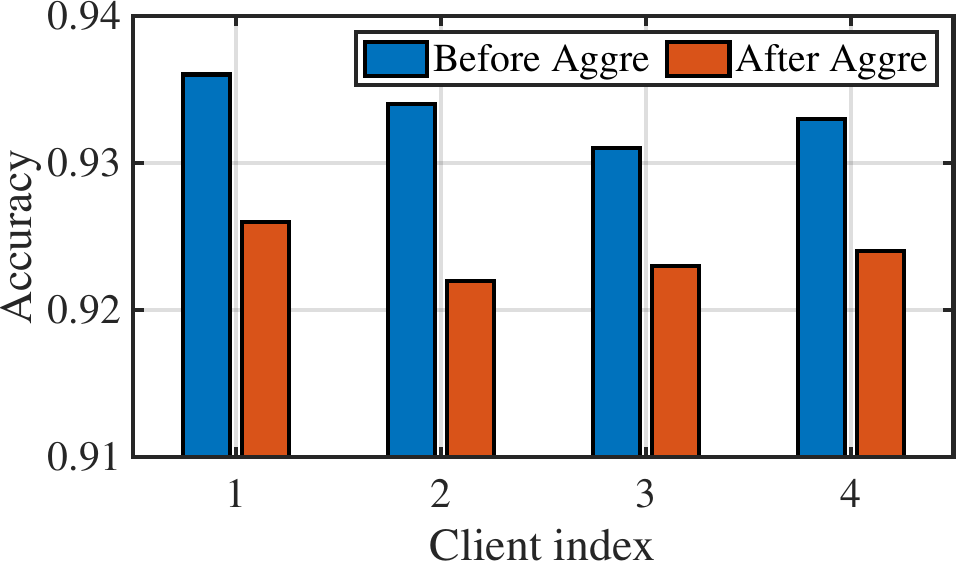}
}
\caption{The activation frequency of experts and comparison of aggregation performance across clients on AGNews dataset.}
\label{fig:mtv_aggregation}
\vspace{-1ex}
\end{figure}

\vspace{-2ex}
\subsection{Model Aggregation Discrepancy} \label{sec:mtv_aggregation}
In conventional federated learning, model aggregation methods~\cite{mcmahan17a, lin2024efficient, zhang2024towards} assume uniform model structures and meaningful updates for all parameters.
\waitrev{However, in MoE-based FL, each client fine-tunes only a sparse expert subsets tailored to its local data distribution through its gating network, resulting in structural discrepancies where many experts remain untouched or never used.
Moreover, clients develop divergent routing preferences as their gating networks tend to favor distinct expert subsets, which leads to inconsistent updates of gating parameters for expert selection in each client.
Directly aggregating such \rev{heterogeneous} model structures introduces destructive interference, 
thus degrading the generalization of the global model.}


\begin{figure*}[t]
\centering
\includegraphics[width=16cm]{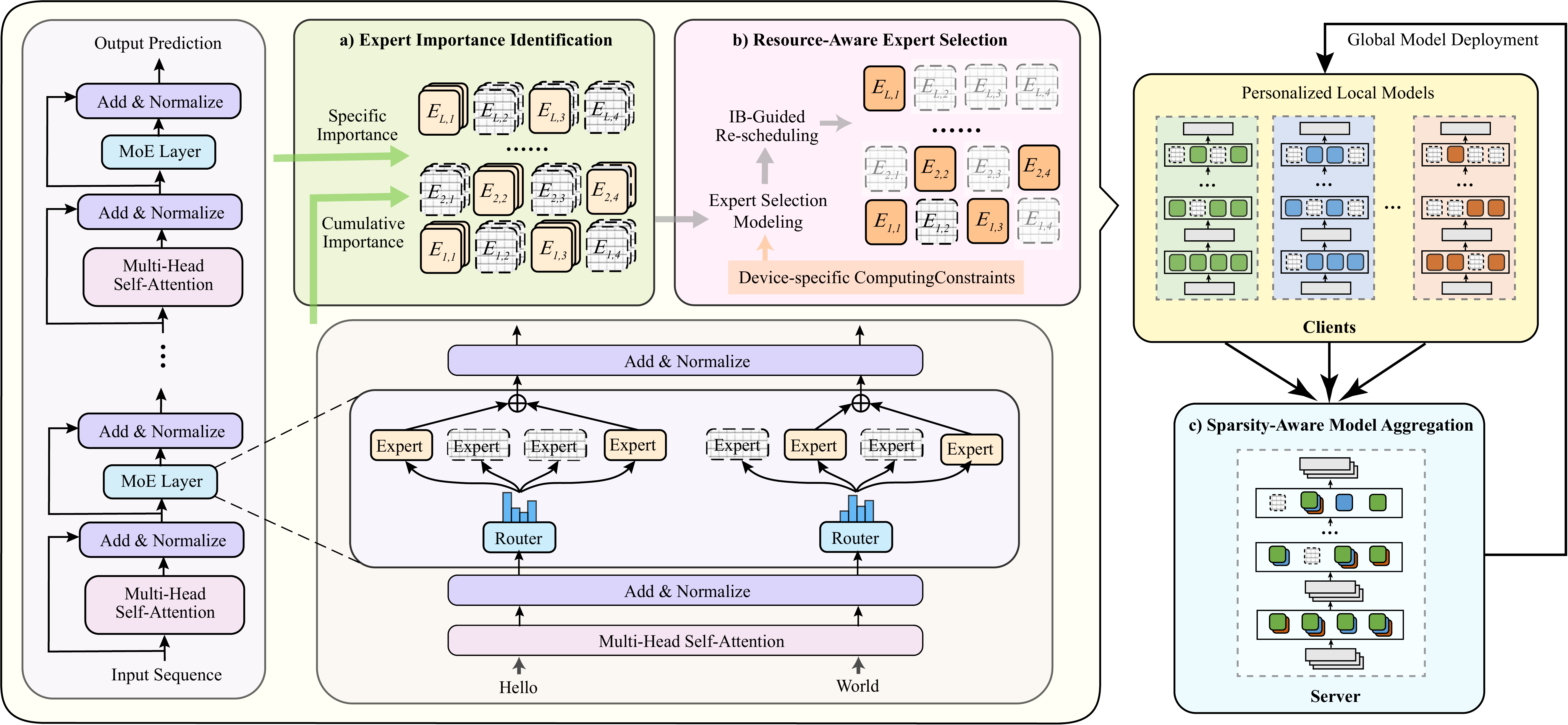}
\caption{Overview of the \name framework. Each client quantifies per-expert importance within a training batch and selects a critical subset of experts for local fine-tuning under device-specific computing constraints. Only the activated experts and their gating networks are uploaded to the server, where a sparsity-aware aggregation strategy is applied to aggregate only active experts and explicitly align routing preferences across clients.}
\label{fig:framwork}
\vspace{-2ex}
\end{figure*}

To empirically demonstrate how model discrepancies impact aggregation performance, we conduct an experiment on AGNews dataset using Switch Transformer with 8 experts,
where each client independently selects and fine-tunes a subset of experts aligned with its local data. It is shown in Fig.~\ref{fig:mtv_aggregation_overlap} that \waitrev{nearly 25\% of the experts are never activated across clients during local fine-tuning, indicating expert discrepancies in parameter updates. In contrast, over 20\% of experts are selected by different clients, highlighting divergent routing preferences across clients.}
Moreover, Fig.~\ref{fig:mtv_aggregation_performance} reveals a notable performance gap in client-specific models before and after standard FedAvg aggregation, highlighting the degraded performance caused by both expert discrepancies and inconsistent gating network updates.
These observations underscore the necessity of sparsity-aware aggregation strategies that explicitly handle the model discrepancies from client-specific routing preferences and expert selections, thereby fully realizing the potential of MoE-based LLM in \rev{federated environments}.

\section{Framework Design}\label{sec:design}
In this section, we introduce \name, an MoE-based FL fine-tuning framework designed for heterogeneous edge computing systems to dynamically select the subset of critical experts.
We will first outline the system overview and then present a detailed description of the \name framework.

\vspace{-2ex}
\subsection{Overview}\label{sec:dsn_overview}
To tackle the above challenges, we propose \name, a resource-efficient FL framework for fine-tuning MoE-based LLMs under computing heterogeneity.
Our design comprises three core components: expert importance identification, resource-aware expert selection, and selective model aggregation.
First, to select appropriate experts for each client, we introduce an expert importance identification scheme that quantifies each expert’s contribution to local fine-tuning performance (Sec.~\ref{sec:dsn_batch_importance}).
Second, to accommodate heterogeneous computing resources across clients during fine-tuning, the resource-aware expert selection module models the expert selection from the information bottleneck perspective (Sec.~\ref{sec:dsn_selection_modeling}) and re-schedules the most critical experts for activation within client-specific computing budgets (Sec.~\ref{sec:dsn_selection_subset}).
Finally, we design a sparsity-aware model aggregation strategy to \waitrev{selectively upload only active experts across \rev{varying model structures} (Sec.~\ref{sec:dsn_aggregation_expert}) and adaptively aggregate the parameters of gating network with importance-weighted contributions  (Sec.~\ref{sec:dsn_aggregation_routing})}, 
enhancing global generalization while preserving critical local updates.

As shown in Fig.~\ref{fig:framwork}, the workflow of \name follows three steps: i) During local fine-tuning, each client identifies the importance of experts based on its local data.
ii) Leveraging the estimated expert importance, clients selectively activate a subset of high-impact experts within their local computing budgets.
iii) After local updates, only the actively trained experts and gating networks are uploaded. The server then performs sparsity-aware model aggregation and distributes the aggregated model to all clients for the next training round.

\vspace{-1ex}
\subsection{System Model} \label{sec:dsn_model}
We consider a federated setting with $C$ clients collaboratively fine-tuning a shared MoE-based LLM under heterogeneous computing resource constraints. Each client $c \in \{1, \dots, C\}$ holds a private dataset $\mathcal{D}_c$, sampled from a distinct data distribution.
The global model consists of a shared gating network $\mathcal{R}$ and \peerrev{an expert set $\mathcal{E} = \{ \mathbf{E}_{1,1}, \dots, \mathbf{E}_{L,S}\}$, where each of the $L$ MoE layers contains $S$ parallel experts.}
For each input token, the gating network computes a routing score and activates a top-$k$ subset of experts per layer ($k \ll S$). This sparse activation allows experts to specialize in local datasets across clients, facilitating personalized model adaptation with reduced computational cost.

During local training round $t$, each client $c$ updates only a selected expert subset $\mathcal{E}_c \subseteq \mathcal{E}$ and gating network $\mathcal{R}_c$ using its local data, constrained by its computing budget $B_c$. Let $\theta_c^t$ denote the local model parameters updated by client $c$ and $\theta^t$ denote the global model. Each client minimizes a local objective
\begin{math}
\mathcal{L}_c(\theta_c^t) = \mathbb{E}_{(x,y) \sim \mathcal{D}_c} \left[ \ell(f(x; \theta_c^t), y) \right],
\end{math}
where $\ell(\cdot)$ is the task-specific loss function and $f(\cdot; \theta_c^t)$ denotes the MoE model's output with parameters $\theta_c^t$.
After local training, clients upload the heterogeneous model parameters $\{\mathcal{E}_c, \mathcal{R}_c\}$ to the server, which then aggregates the received sparse updates to update global model $\theta^{t+1}_{\text{global}}$. 
\peerrev{Our objective is to enable each client to select critical experts tailored to its local data within computing constraints, while collaboratively contributing to a globally shared model that balances accuracy and efficiency.}

\vspace{-2ex}
\subsection{Expert Importance Identification} \label{sec:dsn_batch_importance}
\peerrev{Recalling Sec.~\ref{sec:mtv_expert}, the gating network in MoE is typically optimized for average routing performance across all clients, often failing to accurately prioritize experts most beneficial to each client. This leads to suboptimal expert utilization and degraded fine-tuning performance,
highlighting the need to estimate expert importance for more effective selection.
Therefore, we propose an expert importance identification scheme that quantifies the contribution of each expert to local fine-tuning performance, enabling clients to select a subset of experts that are most relevant to their local data.}

In standard MoE models, the gating network in each layer computes routing score $G_e(x)$ for each expert $e$ based on input token $x$. After softmax normalization, these scores reflect the relative contribution of each expert to the current input. Thus, the top-$k$ experts with the highest scores are selected for fine-tuning as
\begin{equation}
\mathcal{E}_c = \text{Top-}k\left(\{Softmax(\mathbf{G}(x))\}_{e \in \mathcal{E}}\right),
\end{equation}
Finally, \needrev{the output $y$ of this MoE layer} is computed as a weighted sum of the selected experts:
\begin{equation}
y = \sum_{e \in \mathcal{E}_c} G_e(x) \cdot \mathbf{E}_e(x),
\end{equation}
where $\mathbf{E}_e(x)$ represents the output of expert $e$'s network for input token $x$. This routing score $G_e(x)$ serves as an estimate of expert importance.

\peerrev{While this per-token routing mechanism supports dynamic expert selection, it operates on isolated tokens without considering consistent expert activations across samples. However, certain experts are frequently activated in local data and contribute more to specific clients~\cite{lu2024not, guo2021pfl}, failing to capture these cross-sample activations limits model’s ability to learn client-specific expert preferences.
To address this, we quantify expert importance by analyzing expert activations across all tokens within a batch. \rev{As mini-batch processing is standard in LLM training, this incurs negligible overhead while supporting more stable and personalized expert selection for resource-efficient fine-tuning.}}
To the end, we identify experts that are either consistently useful across diverse inputs or particularly important for specific samples as follows.

\begin{itemize}
\item \textbf{Cumulative Importance:}
Experts with high cumulative contributions typically capture \rev{generalizable} features~\cite{muzio2024seer}, providing effective representation across diverse samples.
Prioritizing consistently informative experts minimizes redundant activations, thereby promoting \rev{shared computation} without sacrificing model performance.
As a result, we define cumulative importance of expert $e$ as its \rev{averaged activation} across all samples $x_i$ in the batch of size $B$, which is given by
\begin{equation}
s_b^{cumul}(e) = \frac{1}{B}\sum_{i=1}^{B} G_e(x_i).
\end{equation}
This metric reflects the activation frequency and response strength of expert $e$, serving as a surrogate for its ability to capture label-relevant features across the batch.

\item \textbf{Specific Importance:}
Relying solely on the cumulative contribution across samples fails to capture experts that are infrequently activated but provide critical responses for specific, often hard-to-classify, samples. 
For instance, an expert may contribute substantially to only a few challenging instances, \rev{indicating strong class-specific representational capacity}. Ignoring such experts risks degrading model performance on minority or difficult classes.
To account for this, we introduce the specific importance, which evaluates contribution of experts in individual samples, ensuring that critical responses to challenging inputs are preserved.
Specifically, we use the maximum routing score of each expert across all samples in a mini-batch to capture its \rev{maximum relevance and peak influence} on any single sample, formulated as 
\begin{equation}
s_b^{specific}(e) = \max_{1 \leq i \leq B} G_e(x_i).
\end{equation}
This metric captures the expert’s \rev{peak influence} on single sample, ensuring that low-frequency but high-impact experts are preserved during fine-tuning.
\end{itemize}

Finally, the overall expert importance $s_b(e)$ for expert $e$ in the batch $b$ can be expressed as a weighted combination of its cumulative and specific importance:
\begin{equation} \label{eq:expert_importance}
s_b(e) = \lambda \cdot s_b^{cumul}(e) + (1 - \lambda) \cdot s_b^{specific}(e),
\end{equation}
where $\lambda$ is a hyperparameter that controls the trade-off between activation diversity and specificity. This allows us to strike a balance between generalization and personalization in expert selection, improving both efficiency and effectiveness.

\vspace{-2ex}
\subsection{Resource-aware Expert Selection} \label{sec:dsn_expert_selection}
As explained in Sec.~\ref{sec:mtv_computing}, fine-tuning MoE models on resource-constrained clients is hindered by excessive expert activations across samples. 
Existing studies~\cite{fedus2022switch, dai2024deepseekmoe,lu2024not} primarily select top-$k$ experts per token based on routing scores, overlooking the cumulative computing burden across all tokens within a batch, often exceeding client capabilities.
To address this \rev{bottleneck}, we design a resource-aware expert selection strategy that adaptively limits the number of experts activated per batch according to each client's computing budget. 

\begin{figure}[t]
\vspace{-2ex}
\centering
\subfloat[Different expert numbers \label{fig:dsn_expert_number}]{
\includegraphics[width=0.49\linewidth]{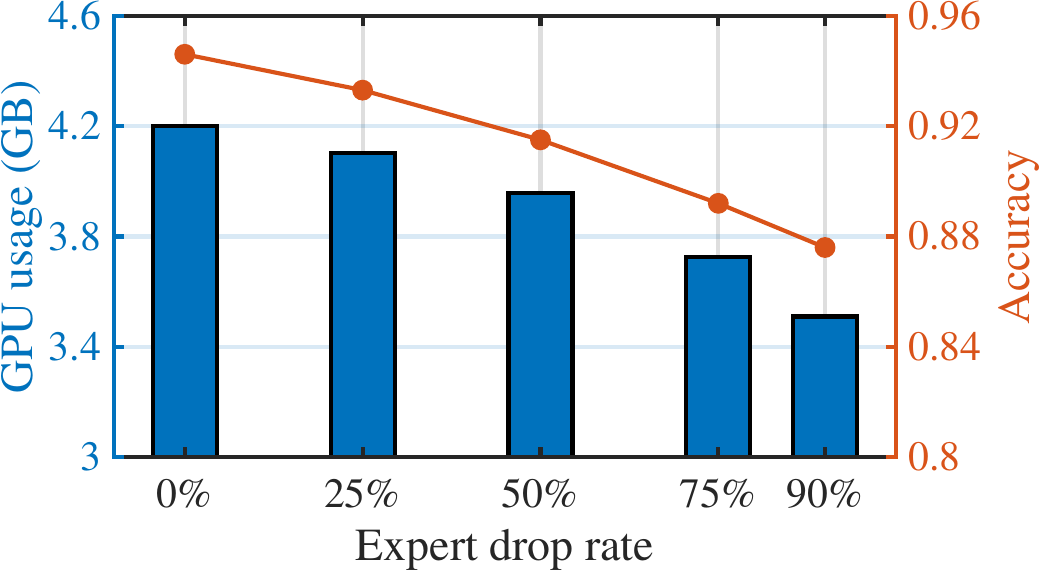}
}
\subfloat[Expert activations \label{fig:dsn_expert_subset}]{
\includegraphics[width=0.49\linewidth]{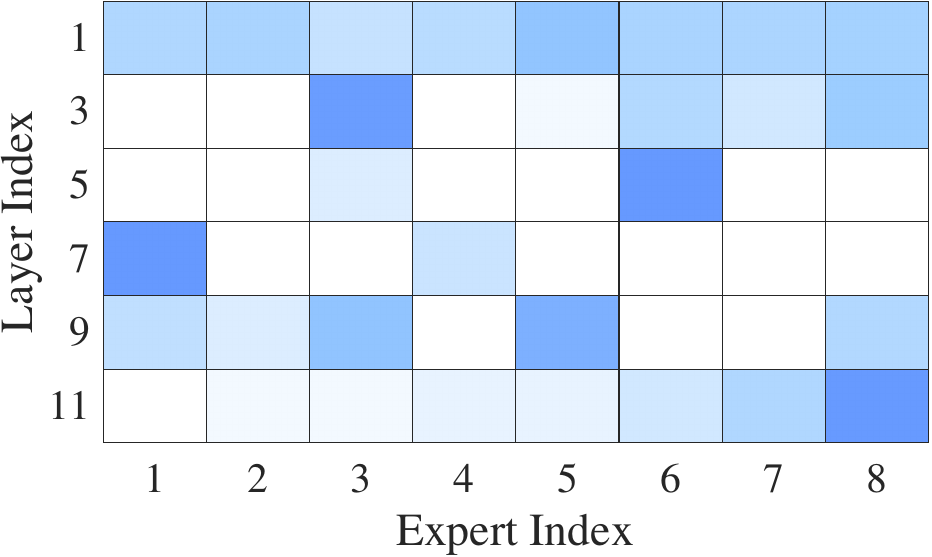}
}
\caption{The impact of various expert dropping rates for fine-tuning the 8-expert Switch Transformer on AGNews dataset.}
\label{fig:dsn_expert}
\vspace{-1ex}
\end{figure}

\subsubsection{\textbf{Expert Selection Modeling}} \label{sec:dsn_selection_modeling}
Achieving efficient fine-tuning on resource-constrained clients necessitates the selective activation of experts within each client’s computing budget.
A straightforward approach is to \peerrev{randomly} drop a number of experts \rev{selected by the gating network.}
However, it is observed in Fig.~\ref{fig:dsn_expert_number} that while reducing the number of experts lowers computational costs, it leads to notable performance degradation.
Fortunately, Fig.~\ref{fig:dsn_expert_subset} reveals that expert selections are highly imbalanced, where a small subset of experts contributes most performance gains, while others offer marginal benefit.
\peerrev{These observations motivate us to narrow the candidate expert set through a contribution-guidance selection strategy, thereby better aligning resource constraints without compromising model performance.}

\peerrev{Inspired by the information bottleneck (IB) principle~\cite{tishby2000information}, the compact subset of experts can be viewed as a compressed representation that retains maximal task-relevant information under limited computational budgets, where the contribution of each expert is quantified by its task-relevant information, and \rev{the minimization of input redundancy encourages sparse expert activation.}}
\peerrev{Therefore, we introduce \rev{an IB-based interpretation} for expert selection modeling, enabling improved explainability and adaptability to heterogeneous client constraints.} 

Specifically, we reinterpret expert activation patterns \rev{over a batch} as a stochastic latent representation $\mathbf{z} = \{z_e, \forall e \in \mathcal{E}\}$,
\needrev{where $z_e \in \{0, 1\}$ indicates whether expert $e$ is activated for a given input $x$.}
From the IB perspective, the expert activation pattern $\mathbf{z}$ serves as a compressed intermediate representation of the input $x$ for predicting the target label $y$. Based on the classical IB formulation~\cite{tishby2000information}, we aim to select a sparse subset of experts by maximizing the following objective:
\begin{equation} \label{eq:ib_objective}
\max_{p(\mathbf{z}|x)} ~ I(\mathbf{z};y) - \beta \cdot I(\mathbf{z};x)
\end{equation}
where \peerrev{$I(\mathbf{z};y)$ measures the task-relevant information preserved in expert activation pattern $\mathbf{z}$, reflecting its contribution to model performance. In contrast, $I(\mathbf{z};x)$ represents the redundancy of input information, serving as a proxy for computational cost.}
The hyperparameter $\beta$ strikes a balance between task-relevant informativeness and resource compression.

Since $I(\mathbf{z};y)$ and $I(\mathbf{z};x)$ are intractable in MoE models, we employ the variational approximation~\cite{alemi2016deep}:
\begin{align}
&I(\mathbf{z};x) \le \mathbb{E}_{x} \left[ KL(p(\mathbf{z}|x)||p(\mathbf{z})) \right] \\
I(\mathbf{z};y) &\ge \mathbb{E}_{(x,y)} \left[ \mathbb{E}_{\mathbf{z} \sim p(\mathbf{z}|x)} (\log q(y|\mathbf{z})) \right] + H(y)
\end{align}
Here, $p(\mathbf{z}|x)$ is the conditional distribution over expert activation given input $x$, \rev{$p(\mathbf{z}) \sim \mathcal{N}(0,I)$ is the prior, and $q(y|\mathbf{z})$ is a variational decoder that approximates the label likelihood.}
\peerrev{As $H(y)$ is independent of expert behavior, it can be omitted for importance comparison (i.e., in the formulated optimization).
Under the principle of optimal information compression, we approximate the intractable joint optimization in the original IB objective (Eqn.~\eqref{eq:ib_objective}) by independently evaluating the IB contribution of each expert within a batch $b$.} For each expert $e$, its IB contribution is defined as
\begin{equation}
\begin{split}
I_b(e) &= I(z_e;y) - \beta I(z_e;x) \ge \mathbb{E}_{z_e \sim p(z_e|x)} \left[ \log q(y|z_e) \right] \\
&+ H(y) - \beta \cdot KL(p(z_e|x)||p(z_e))
\end{split}
\end{equation}
where $I_b(e)$ denotes the informativeness of expert $e$. Expert $e$ is considered more critical when $I_b(e)$ is higher, as it suggests greater task relevance and a more compact representation.

In MoE frameworks, expert activations $z_e$ are not explicitly modeled but are implicitly determined by the gating network. 
Since the routing score $G_e(x)$ reflect how input $x$ are assigned to expert $e$, the activation distribution $p(z_e|x)$ can be naturally approximated by $G_e(x)$.
Consequently, the marginal distribution $p(z_e)$ can be estimated as \rev{its averaged routing probability} over a batch of size $B$:
\begin{equation}
p(z_e|x) = G_e(x),~p(z_e) = \frac{1}{B} \sum_{j=1}^{B} G_e(x_j),
\end{equation}

Therefore, the approximation of $I(z_e;x)$ and $I(z_e;y)$ can be  represented directly from routing behavior as
\begin{align}
K&L(p(z_e|x)||p(z_e)) = \frac{1}{B} \sum_{i=1}^B G_e(x_i) \log \frac{G_e(x_i)}{p(z_e)}, \\
&\mathbb{E}_{z_e \sim p(z_e|x)} \left[ \log q(y|z_e) \right] = p(z_e|x) \cdot \log q(y|z_e).
\end{align}

However, the label likelihood conditioned on the expert activation $q(y|z_e)$ is not explicitly available. To address this, we estimate $I(z_e;y)$ by analyzing the routing behavior of expert $e$.
The key insight is that if $G_e(x)$ is consistently high for a group of inputs $x$ \rev{sharing the same label $y_i$}, \waitrev{then expert $e$ likely captures representative and discriminative patterns relevant to $y_i$.}
\needrev{Under this observation, routing behavior itself indirectly reflects mutual information:}
\begin{equation}
I(z_e;y) \approx f(\left \{G_e(x)\right \}_{i=1}^B) 
\end{equation}

To facilitate efficient and interpretable estimation, we instead derive a tractable approximation that \rev{decomposes $I(z_e;y)$ into two interpretable components: cumulative importance $s_b^{cumul}(e)$ and specific importance $s_b^{specific}(e)$, as defined in Eqn.~\eqref{eq:expert_importance}.}
%
To this end, the IB contribution $I_b(e)$ for expert $e$ over the entire batch $b$ can be expressed as
\begin{equation} \label{eq:expert_information}
\begin{split}
I_b(e) = s_b(e) - \beta \cdot \frac{1}{B} \sum_{i=1}^B G_e(x_i) \log \frac{G_e(x_i)}{p(z_e)}
\end{split}
\end{equation}
A higher $I_b(e)$ indicates that expert $e$ is both informative and compact, making it a strong candidate for selective fine-tuning under computing resource constraints.

\begin{figure}[t!]
\centering
\includegraphics[width=7.2cm]{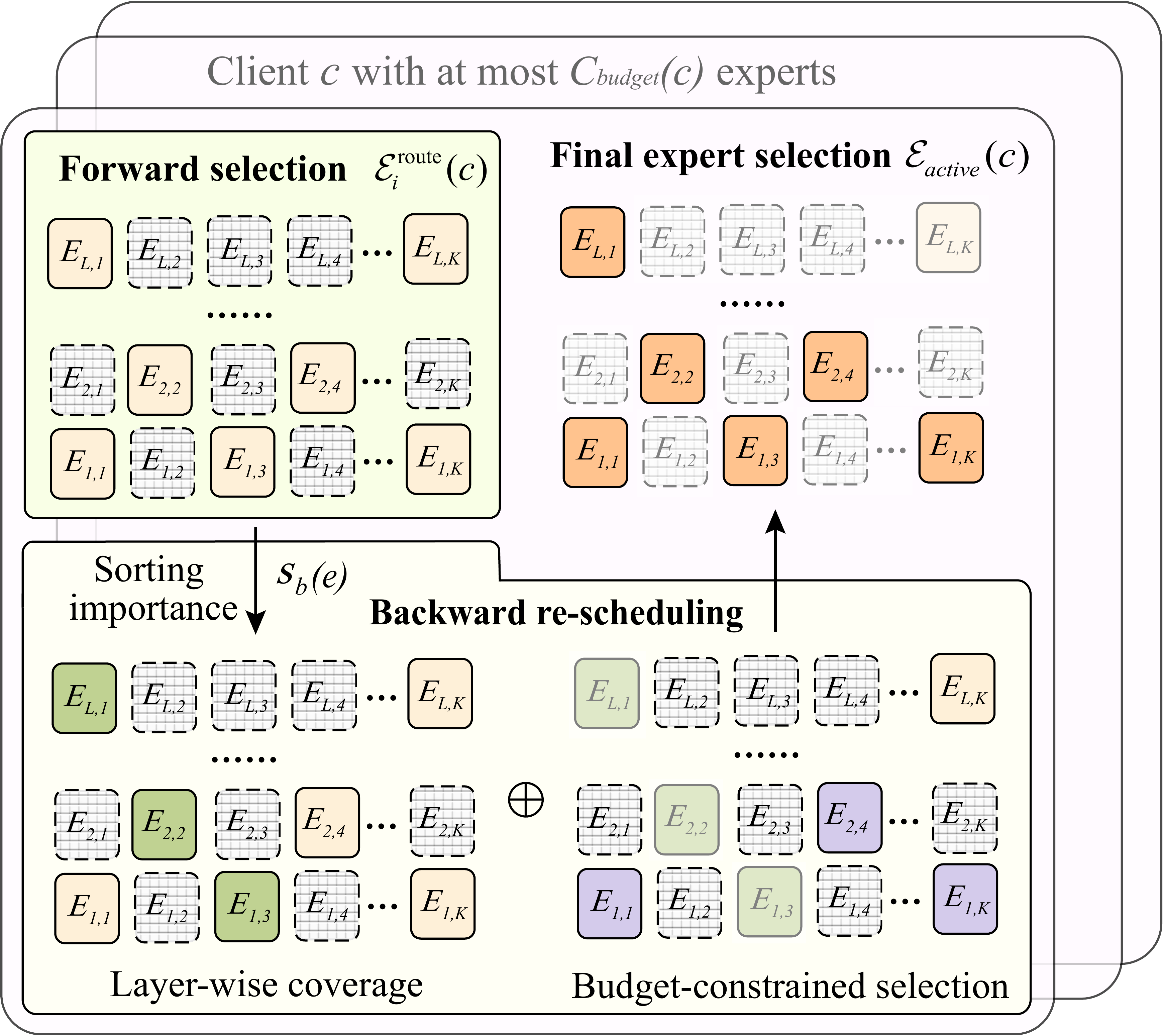}
\caption{The resource-aware expert selection under computing constraints, where expert activation is re-scheduled during the backward pass based on IB-guided importance ranking.}
\label{fig:selection}
\vspace{-2ex}
\end{figure}

\subsubsection{\textbf{IB-Guided Expert Re-scheduling}} \label{sec:dsn_selection_subset}
Given the expert contribution $I_b(e)$ for all experts $e \in \{\mathbf{E}_1, \dots, \mathbf{E}_S\}$ in client $c$'s model, \peerrev{we aim to select a subset $\mathcal{E}_{\text{active}}(c)$ for each client $c$ of at most $C_{\text{budget}}(c)$ experts to be activated across the batch, where the most informative experts are prioritized for activation.}
Let $\mathcal{E}_{\text{union}}(c) = \bigcup_{i=1}^{B} \mathcal{E}_i^{\text{route}}(c)$ denote the union of selected experts for all samples in batch $b$, where gating network selects top-$k$ experts $\mathcal{E}_i^{\text{route}}$ for each sample. 
Under the computing budget $C_{\text{budget}}(c)$ for client $c$ (i.e., the maximum number of experts can be activated per batch), we solve the following selection problem, guided by the optimization objective in Eqn.~\eqref{eq:ib_objective}, as
\begin{equation} \label{eq:expert_selection}
\mathcal{E}_{\text{active}}(c) = 
\arg \max_{\substack{
\mathcal{E}_{\text{active}}(c) \subseteq \mathcal{E}_{\text{union}}(c) \\
|\mathcal{E}_{\text{active}}(c)| \leq C_{\text{budget}}(c)
}} 
\sum_{e \in \mathcal{E}_{\text{active}}(c)} I_b(e)
\end{equation}

To achieve this, we re-schedule forward-selected experts to activate only critical ones during backward propagation, as shown in Fig.~\ref{fig:selection}.
In the forward pass, for each sample $x_i$ in the batch, the gating network selects its top-$k$ experts $\mathcal{E}_i^{\text{route}}(c)$ based on the routing scores, preserving the original sample-level routing decisions.
In the backward pass, we re-schedule expert activation across the batch to satisfy the computing constraint using the following resource-aware procedure:


\begin{itemize}

    \item \textbf{Importance-based sorting:} We sort all $e \in \mathcal{E}_{\text{union}}(c)$ in descending order based on their importance scores $s_b(e)$.
    \item \textbf{Layer-wise coverage:} To ensure all MoE layers remain functional, we first select the most important expert in each layer.
    \item \textbf{Budget-constrained selection:} With the remaining budget, we continue selecting the highest-scoring experts regardless of their layer, until the total number of activated experts reaches the budget limit $C_{\text{budget}}(c)$.
\end{itemize}

During backward propagation, only the selected experts in $\mathcal{E}_{\text{active}}$ participate in gradient computation and parameter updates. Gradients for all other inactive experts are masked, thus reducing computational cost without degrading model performance.
The expert selection process is refreshed in each training round, ensuring dynamic adaptation to local data. The model is optimized using standard task loss functions (e.g., cross-entropy) without introducing any additional regularization during the expert selection phase.

\vspace{-2ex}
\subsection{Sparsity-aware Model Aggregation} \label{sec:dsn_aggregation}
\begin{figure}[t!]
\centering
\includegraphics[width=8cm]{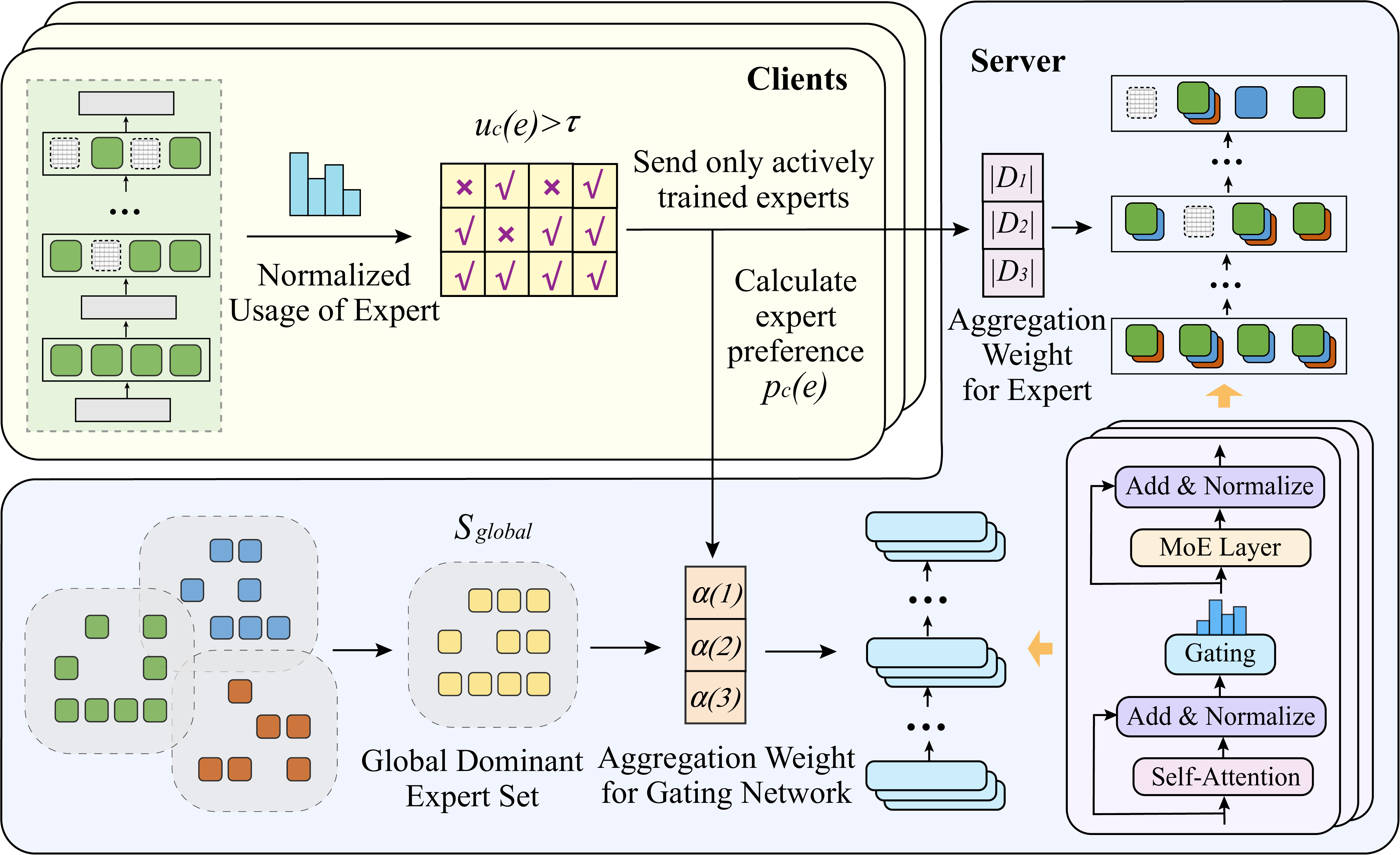}
\caption{The sparsity-aware model aggregation. Active experts are selectively aggregated, and gating parameters are importance-weighted to align routing behaviors across clients.}
\label{fig:aggregation}
\vspace{-2ex}
\end{figure}

\waitrev{As discussed in Sec.~\ref{sec:mtv_aggregation}, client-specific constraints lead to distinct routing preferences and personalized expert selections, which introduces destructive interference and severely undermines the performance of the global model.
These discrepancies are further exacerbated by heterogeneous computing resources, as the resource-aware expert selection activates only important experts on each client.
To address this issue, as shown in Fig.~\ref{fig:aggregation}, we develop a sparsity-aware model aggregation strategy that updates only the experts actively trained on clients, while aggregating gating parameters with importance-weighted contributions to ensure the global router properly captures client-specific routing behaviors.}
We summarize the proposed MoE-based LLM fine-tuning via FL in Algorithm~\ref{alg:name}.

\subsubsection{\textbf{Selective Expert Aggregation}} \label{sec:dsn_aggregation_expert}
In MoE-based federated learning, each client updates only a subset of experts based on its local gating network, leaving many experts partially trained or entirely untouched with random initialization.
However, standard aggregation such as FedAvg~\cite{mcmahan17a} assumes a consistent model structure with meaningful updates for all parameters among clients. This mismatch causes \rev{destructive interference} when aggregating irrelevant or stale expert weights, and thus degrading the generalization of the global model.

To address \rev{discrepancies} in partial expert updates, we propose a selective expert aggregation strategy.
Instead of applying direct aggregation across all experts indiscriminately, we only update experts that have been actively trained on a client.
Denote $u_{c}(e)$ as the \needrev{normalized usage} of expert $e$ on client $c$ over local samples $D_{c}$, which is estimated as
\begin{equation} \label{eq:active_expert}
u_{c}(e) = \frac{1}{\left | D_{c} \right |} \sum_{x_i \in D_{c}} G_e(x_i).
\end{equation}

To filter out unreliable or inactive expert updates, we introduce a usage threshold $\tau$, such that each client uploads only the experts with usage $u_{c}(e) \ge \tau$. 
\peerrev{The server then keeps the inactive experts unchanged and only aggregates these active experts as follows}
\begin{equation} \label{eq:expert_aggregation}
E_e^{t+1} = \sum_{c=1}^C \frac{\vert D_{c} \vert {\bf 1}(u_j(e)\ge\tau)}{\sum_{j=1}^C \vert D_j\vert {\bf 1}(u_j(e)\ge\tau)}  E_e^t(c) \cdot \mathbf{1}(u_c(e) \ge \tau),
\end{equation}
where $E_e^t(c)$ represent the parameters of expert $e$ on the client $c$ after local fine-tuning in round $t$. 
\waitrev{The indicator function $\mathbf{1}(\cdot)$ ensures that expert $e$ from a client contributes to aggregation only when it has been actively trained; otherwise, it is excluded if it remains unused or undertrained.}
This strategy prevents updates from clients with insufficient interactions with experts, thereby \rev{improving parameter consistency and preserving expert specialization.}



\subsubsection{\textbf{Importance-Weighted Gating Aggregation}} \label{sec:dsn_aggregation_routing}
Apart from partial expert updates, clients also develop divergent routing preferences as their gating networks tend to prioritize distinct expert subsets. 
\waitrev{This divergence yields inconsistent values of gating parameters across clients, reflecting distinct expert activation patterns.
Directly aggregating gating parameters without accounting for client-specific routing preferences dilutes the locally learned expert selection patterns, leading to ambiguous routing decisions. 
As a result, the aggregated gating network struggles to adapt expert selection to local data, undermining training stability and thus hindering model convergence.}


To mitigate this inconsistency, we propose an importance-weighted gating aggregation strategy that explicitly incorporates expert preference and routing consistency. 
The key idea is to adjust each client’s contribution to the global gating network according to its expert preferences and their consistency with the global distribution.
Clients with higher contributions are assigned larger aggregation weights, ensuring that the global gating network is not distorted by divergent routing behaviors.




To quantify the consistency of routing behaviors across clients, we focus on the dominant experts that capture the routing preferences of each client.
For each client $c$, the dominant expert subset is determined by the actively trained experts under local data, which contribute most to routing decisions. 
Given the normalized usage $u_c(e)$ of expert $e$ in Eqn.~\eqref{eq:active_expert}, the dominant expert subsets of client $c$ can be expressed as $\mathcal{S}_c = \{e|u_c(e) \ge \tau \}$, and \needrev{the global dominant expert set is $\mathcal{S}_{\text{global}} = \cup \mathcal{S}_c$.} 
The routing consistency $r(c)$ of client $c$ is then quantified as the proportion of overlap between its dominant experts and the global dominant expert set:
\begin{equation} \label{eq:routing_consistency}
r(c) = \frac{|\mathcal{S}_c \cap \mathcal{S}_{\text{global}}|}{|\mathcal{S}_{\text{global}}|},
\end{equation}
where a higher value indicates more aligned and reliable routing behavior for aggregation. 

\waitrev{While dominant expert overlap captures whether clients focus on similar subsets of experts, it fails to account for the relative significance of those experts, which may differ substantially across local data distributions. To further account for the contribution of individual experts, we incorporate expert importance into the aggregation weights.}
As the expert importance score $s_b(e)$ in Eqn.~\eqref{eq:expert_importance} provides an estimate of expert importance under local data distribution, it is adopted as the expert preference to ensure that experts with greater influence on local performance play a more decisive role in the aggregation of the global gating network.
By calculating the expert importance in dominant expert subsets, the preference of expert $e$ for client $c$ is given by
\begin{equation} \label{eq:expert_preference}
p_c(e) = u_c(e) \cdot s_b^c(e), \forall e \in S_c.
\end{equation} 
 
Finally, the aggregation weight of client $c$ is defined as \rev{the joint consideration of its routing consistency $r(c)$ and expert preference $p_c(e)$:}
\begin{equation}
\alpha(c) = \frac{\sum_{e \in \mathcal{S}_c} p_c(e)}{|\mathcal{S}_{\text{global}}|},
\end{equation}
which emphasizes experts that are both commonly selected and locally important.
The global gating network parameters are then updated through importance-weighted averaging:
\begin{equation} \label{eq:gating_aggregation}
G_{\text{gating}}^{t+1} = \sum_{c=1}^C \alpha(c) \cdot G_c^t,
\end{equation}
where $G_c^t$ denotes the parameters of gating network in client $c$ at the $t$-th training round. 
This gating aggregation strategy ensures that clients with experts that are both frequently activated and consistently utilized have greater impact on the aggregation, thereby improving stability and convergence of MoE-based FL training.


\RestyleAlgo{ruled}
\LinesNumbered
\begin{algorithm}
\caption{Federated MoE Fine-tuning with Resource-Aware Expert Selection}
\label{alg:name}
\setstretch{1.0}
\small
\textbf{Input:} Total training rounds $T$, local epochs $E$, Computing budget $C_{\text{budget}}$\\
Initialize global model $\theta^{0}$ for all experts $e$\;
\For{each round $t = 1$ to $T$}{
    \For{each client $c$}{
        Download global model $\theta_{\text{global}}^{t}$ and initialize local model $\theta_c \gets \theta_{\text{global}}^{t}$\;
        \For{each local epoch $e = 1$ to $E$}{
            \For{each mini-batch $\mathcal{B}$}{
                Compute routing scores $G_e(x_i)$ for each sample $x_i$ and expert $e$\;
                Compute expert importance using Eqn.~\eqref{eq:expert_information}\;
                Select experts  $\mathcal{E}_{\text{active}}$ under $C_{\text{budget}}(c)$ via Eqn.~\eqref{eq:expert_selection}\;
                Backward only on selected experts\;
                Update $\theta_c$ by SGD\;
            }
        }
        Calculate each expert's $u_c(e)$ with Eqn.~\eqref{eq:active_expert} and filter inactive experts\;
        Calculate the sum of expert preference $\sum_{e \in \mathcal{S}_c} p_c(e)$ over the dominant expert subsets\;
        Upload active experts, other model updates, and the sum of expert preference to server\;
    }
    Server aggregates experts and gating networks with Eqn.~\eqref{eq:expert_aggregation} and Eqn.~\eqref{eq:gating_aggregation}\;
}
\textbf{Output:} Final global model $\theta_{\text{global}}^{T}$\\
\end{algorithm}

\vspace{-2ex}
\section{Simulation Setup} \label{sec:simulation_simu_setup}
In this section, we demonstrate the detailed simulation setup to evaluate our \name for heterogeneous MoE-based LLM fine-tuning using Switch Transformer with 64 experts per layer and DeepSeek-MoE-16 models. The performance of \name is evaluated against several baselines using carefully selected hyper-parameters to ensure a fair comparison.

\subsubsection{\textbf{Model and dataset}}
For our experiments, we employ two representative MoE-based LLMs, Switch Transformer~\cite{fedus2022switch} and DeepSeek-MoE-16B~\cite{dai2024deepseekmoe}. 
Switch Transformer adopts a top-1 gating mechanism for 64 experts per layer, enabling efficient scaling to over 395B model parameters while activating only 11B parameters. DeepSeek-MoE-16B leverages a more expressive backbone with a top-2-of-64 gating, achieving comparative performance of LLaMA2-7B model. 
To evaluate model performance, we report test accuracy for \rev{converged global model} on excessive widely used datasets covering tasks from basic semantic classification to advanced knowledge reasoning:
(1)~AGNews~\cite{zhang2015character}: Four-class topic classification of news titles and descriptions across World, Sports, Business, and Sci/Tech domains.
(2)~PIQA~\cite{bisk2020piqa}: A binary-choice dataset for physical commonsense reasoning. Given a short scenario, the model selects the more plausible of two solutions.
(3)~HellaSwag~\cite{zellers2019hellaswag}: A four-way multiple-choice benchmark for commonsense reasoning. Each instance requires selecting the most plausible continuation of a given context. The data is adversarially filtered to discourage superficial pattern matching.
(4)~MMLU~\cite{hendrycks2020measuring}: A 57-task benchmark covering diverse academic and professional domains. Each task contains four-choice questions designed to test multi-domain reasoning.

\subsubsection{\textbf{Baselines}}
To investigate the advantages of our \name framework, we compare it with the following baselines:
\begin{itemize}
  \item \textbf{FedAvg~\cite{mcmahan17a}} aggregates model weights via weighted averaging, serving as a standard baseline for FL optimization.
  \item \textbf{PFL-MoE~\cite{guo2021pfl}} integrates MoE with personalized federated learning by allowing each client to train a subset of shared experts based on local data, enabling client-specific expert selection from globally shared knowledge. 
  \item \textbf{FedMoE~\cite{mei2024fedmoe}} reduces the per-client computing burden by identifying a suboptimal submodel during preliminary fine-tuning, limiting expert activation to a smaller subset.
  \item \textbf{SEER-MoE~\cite{muzio2024seer}} enhances the efficiency of sparse expert activation in MoE models by pruning the total number of experts with gating regularization, optimizing computation through better utilization of expert resources.
\end{itemize}

\begin{figure}[t!]
\centering
\includegraphics[width=8.2cm]{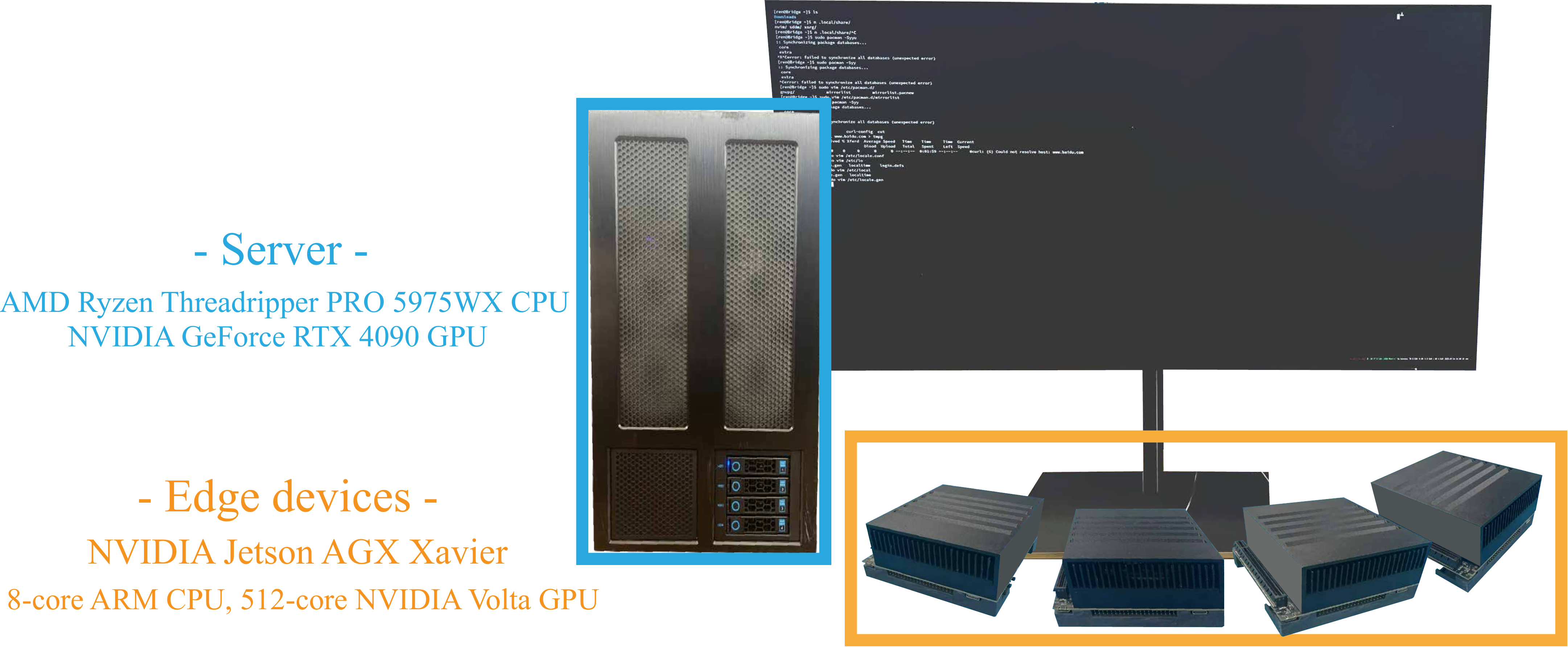}
\vspace{0.05cm}
\caption{Our \name prototype and testbed.}
\label{fig:implementation}
\vspace{-2ex}
\end{figure}

\subsubsection{\textbf{Hyper-parameters}} 
In the experiments, we implement \name prototype for MoE-based LLM instruction fine-tuning using Switch-base-64 and DeepSeek-MoE-16B as pre-trained backbone models, with 8-rank QLoRA applied for loading DeepSeek-MoE-16B model.
\peerrev{As illustrated in Fig.~\ref{fig:implementation}, our distributed system consists of a central server equipped with NVIDIA GeForce RTX 4090 GPUs and $C=4$ clients running in synchronized mode, each implemented on NVIDIA Jetson AGX Xavier kits.}
Each client uses a learning rate of 0.0001 and the batch size for local fine-tuning is set to $B=8$. We set the hyperparameters as $\lambda=0.9$, $\beta=0.1$, and $\tau=0.05$.
To simulate computing heterogeneity, we constrain each client's maximum computing resource available for fine-tuning \needrev{by sampling uniformly from 12\!~GB to 32\!~GB. Since the fine-tuning of Switch-base-64 and DeepSeek-MoE-16B requires at least 22\!~GB, approximately 50\% of clients have to select experts in order to participate in fine-tuning.}

\begin{figure}[t]
\vspace{-2ex}
\centering
\subfloat[AGNews \label{fig:agnews_acc}]{
\includegraphics[width=0.49\linewidth]{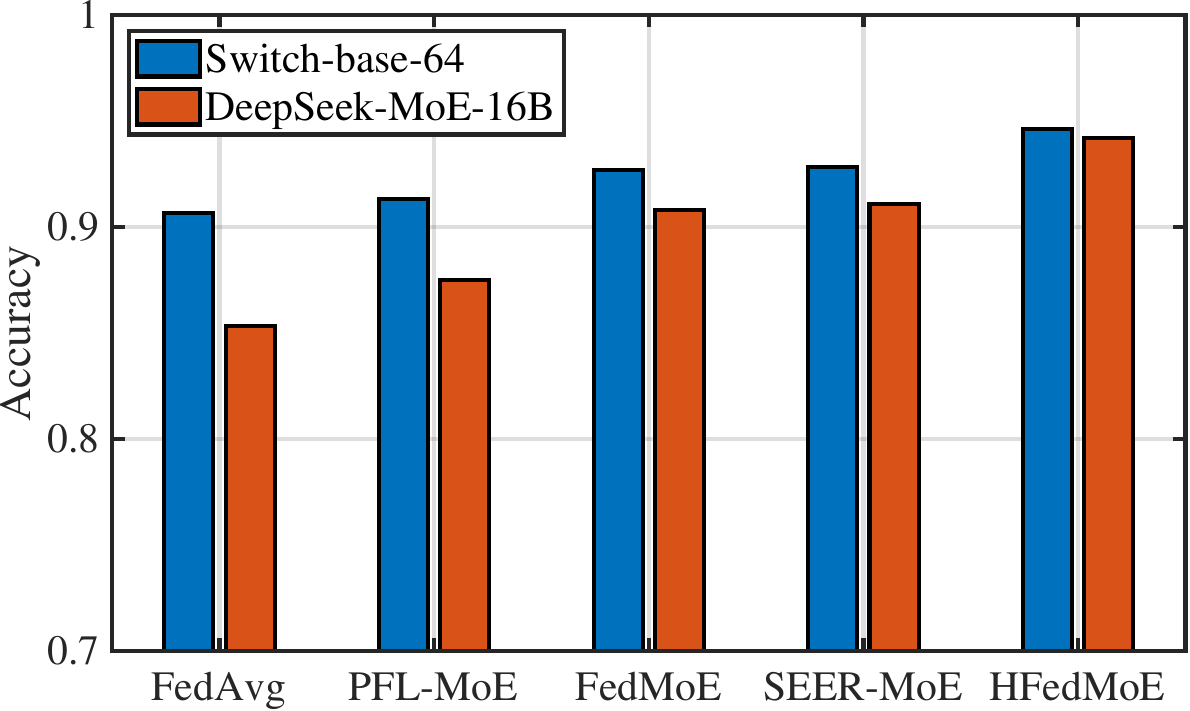}
}
\subfloat[PIQA \label{fig:piqa_acc}]{
\includegraphics[width=0.49\linewidth]{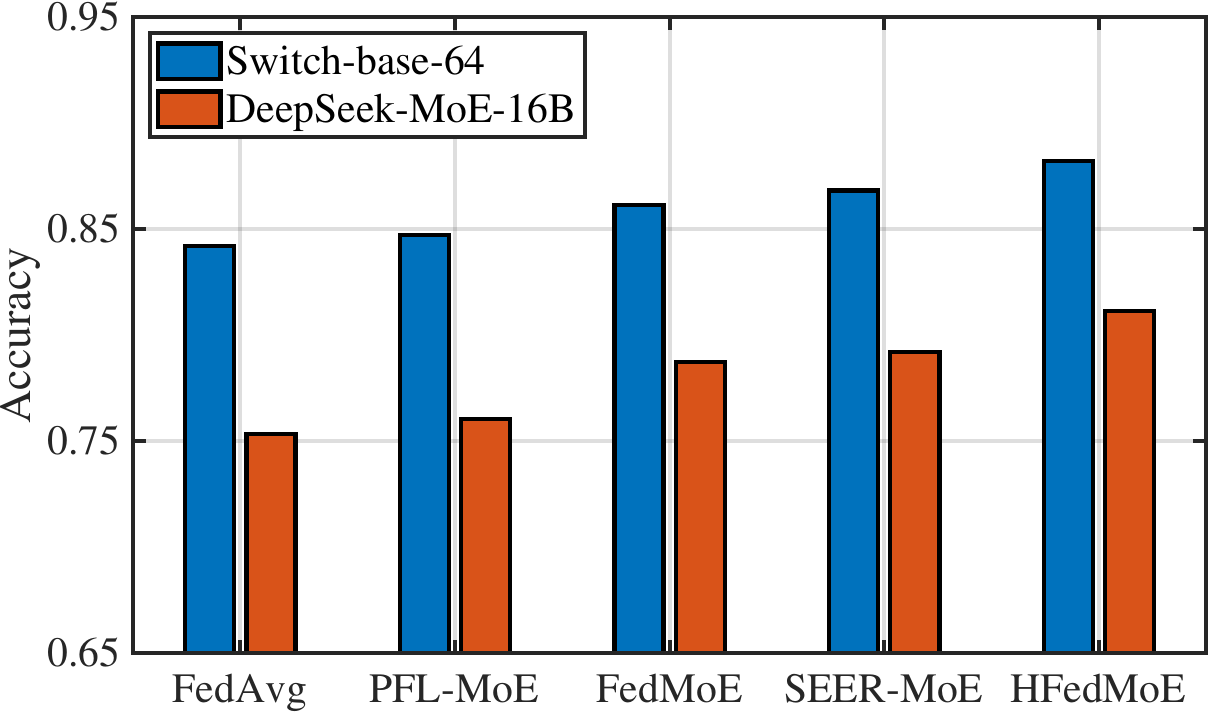}
} \\
\vspace{-2ex}
\subfloat[HellaSwag \label{fig:hellaswag_acc}]{
\includegraphics[width=0.49\linewidth]{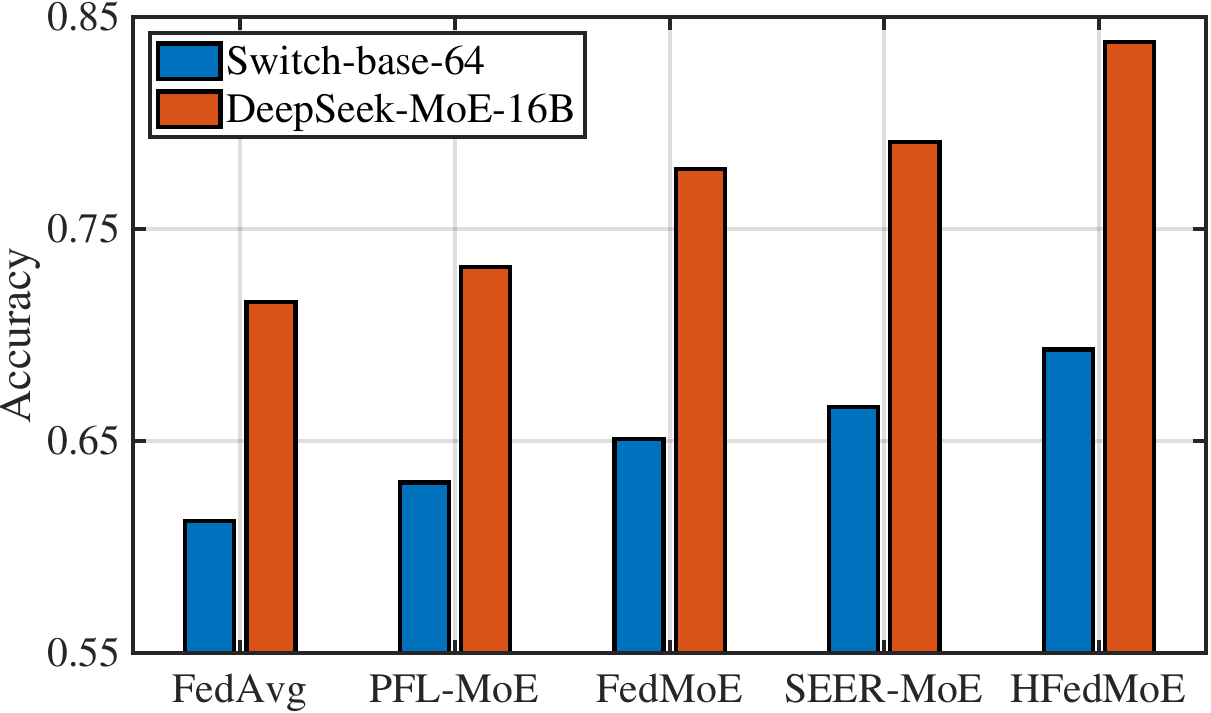}
}
\subfloat[MMLU \label{fig:mmlu_acc}]{
\includegraphics[width=0.49\linewidth]{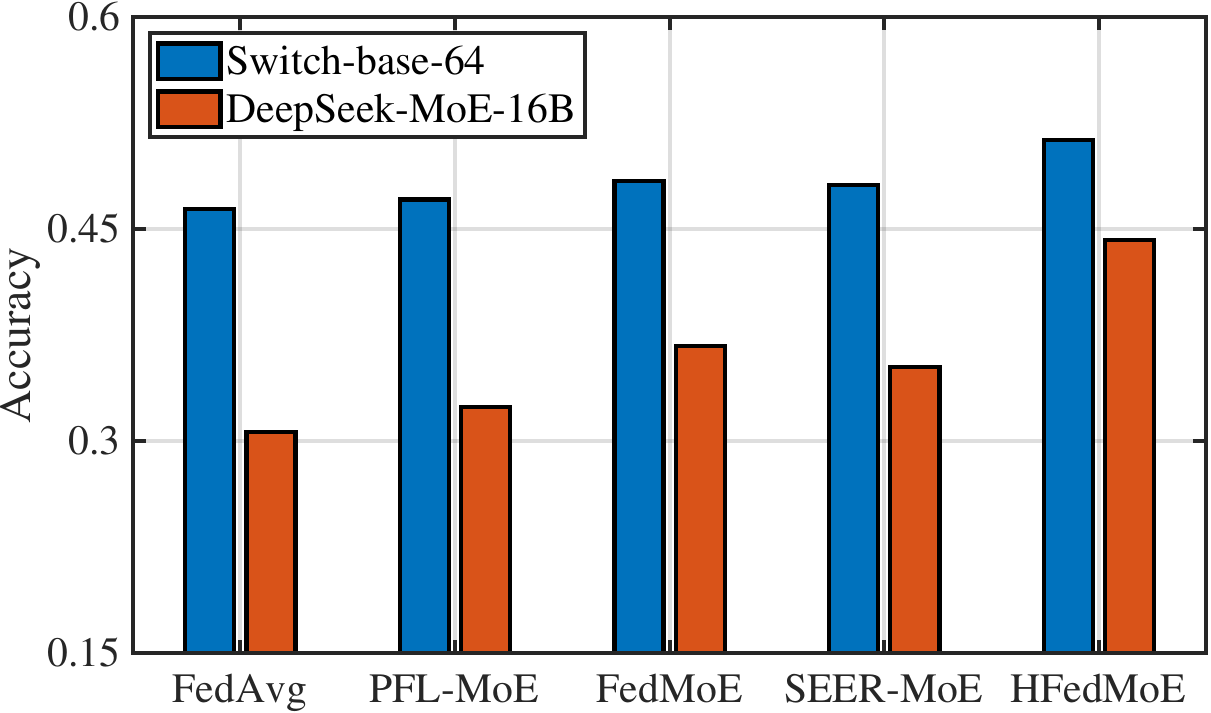}
}
\vspace{-1ex}
\caption{The test accuracy across benchmarks on four datasets with Switch-base-64 and DeepSeek-MoE-16B models under computing heterogeneity.}
\label{fig:overall_accuracy}
\vspace{-1ex}
\end{figure}

\section{{Performance Evaluation}} \label{sec:simulation}
In this section, we evaluate the overall performance and the computing efficiency of \name against various benchmarks. We further investigate the impact of different hyper-parameter settings within our \name framework.
The contributions of each meticulously designed component in \name are also analyzed to illustrate their individual roles in the proposed framework.

\vspace{-1ex}
\subsection{The Overall Performance}
\subsubsection{\textbf{The test accuracy}}
Fig.~\ref{fig:overall_accuracy} presents the test accuracy of \name and other benchmarks on four datasets under computing heterogeneity. 
\name demonstrates superior performance across diverse benchmarks, achieving test accuracy over \needrev{94\%, 81\%, 72\%, and 45\%} on AGNews, PIQA, HellaSwag, and MMLU, respectively. 
\peerrev{Compared to federated MoE fine-tuning, such as FedMoE and PFL-MoE, \name greatly improves model performance
due to the proposed sparsity-aware model aggregation strategy, which aligns client-specific routing preference and aggregates heterogeneous MoE submodels with selective expert activation.}
While FedMoE and SEER-MoE also reduce the number of activated experts, their performance degrades significantly 
due to the lack of accurate expert importance identification and the inability to handle concurrent expert activations during expert selection, leading to suboptimal model updates and degraded generalization.
In contrast, by identifying expert importance across samples 
and dynamically selecting critical experts to align with each client's computing budgets, \name enforces resource-aware expert selection to enable efficient local fine-tuning without sacrificing performance.

\subsubsection{\textbf{\rev{The Convergence Performance}}}
\begin{figure}[t]
\centering
\subfloat[AGNews \label{fig:convergence_agnews}]{
\includegraphics[width=0.49\linewidth]{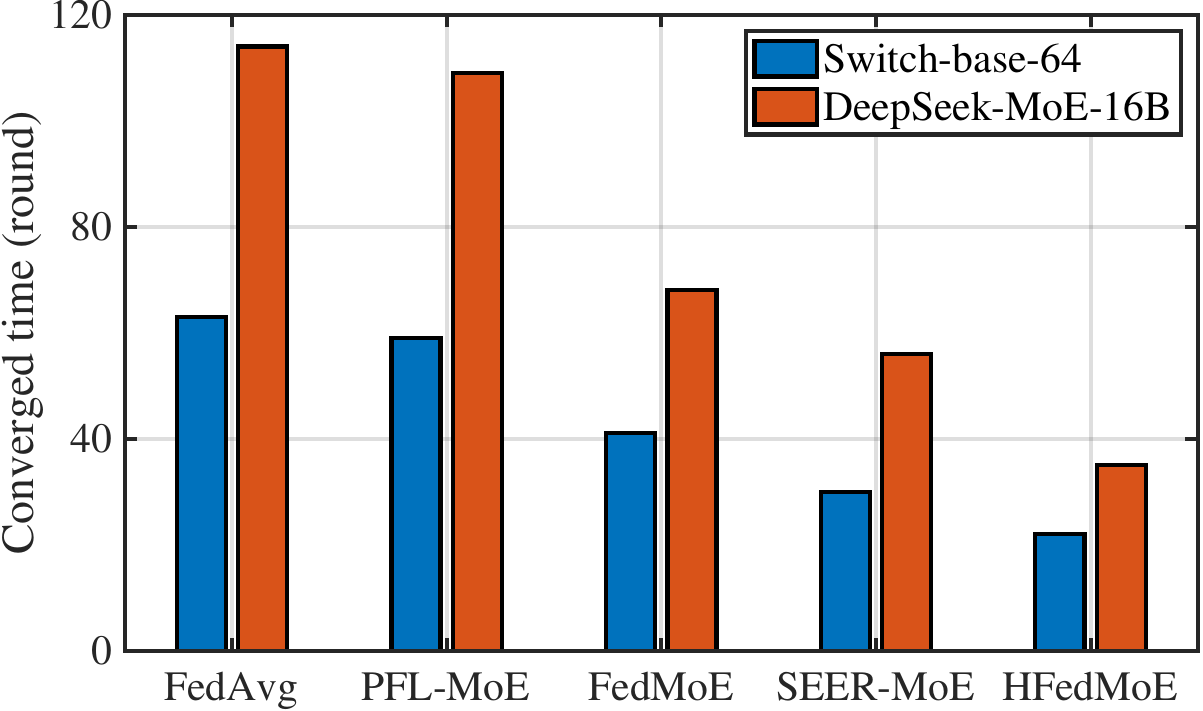}
}
\subfloat[MMLU \label{fig:convergence_mmlu}]{
\includegraphics[width=0.49\linewidth]{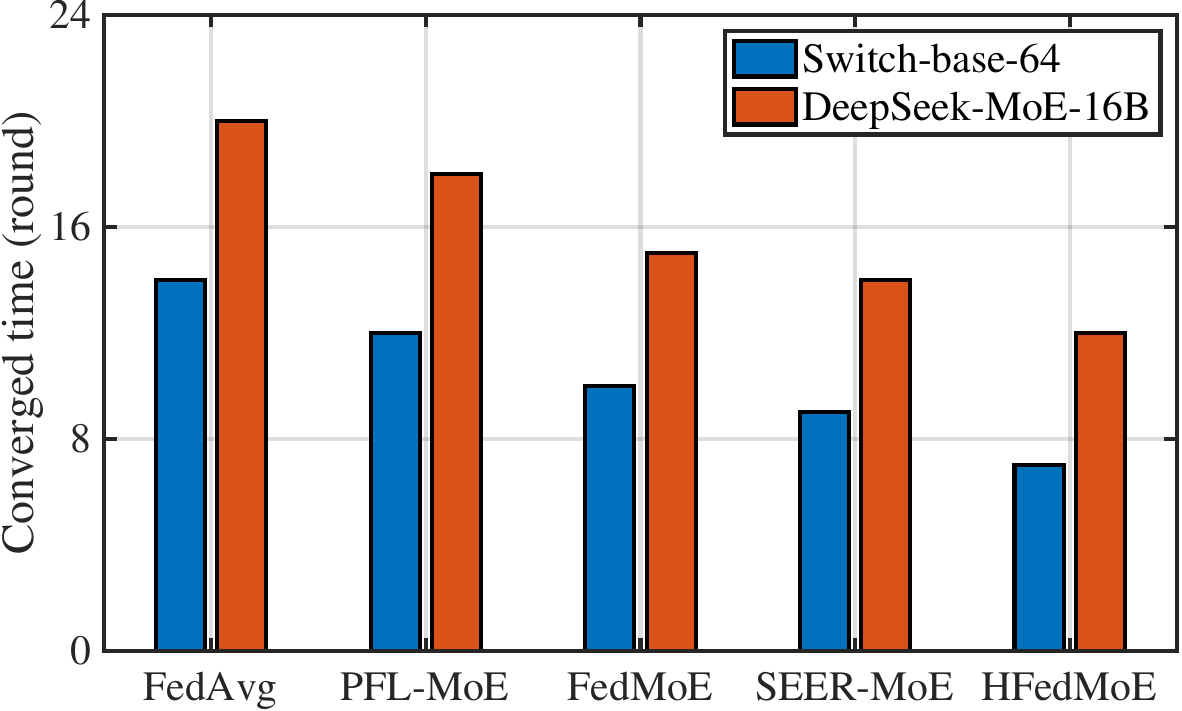}
}
\vspace{-1ex}
\caption{The convergence time on the AGNews and MMLU dataset under computing heterogeneity.}
\label{fig:convergence_time}
\vspace{-3ex}
\end{figure}

Fig.~\ref{fig:convergence_time} compares the convergence time of \name and other four benchmarks on AGNews (classification) and MMLU (reasoning) tasks using Switch-base-64 and DeepSeek-MoE-16B models \needrev{under heterogeneous computing constraints}. 
Across both datasets and models, \name consistently achieves fastest convergence, yielding over \needrev{x1.4} and \needrev{x1.6} speedups on Switch-base-64 and DeepSeek-MoE-16B models, respectively. 
This improvement arises from \name's resource-aware expert selection, which mitigates local fine-tuning failures caused by computational overload while preserving the most informative expert activations for fine-tuning.
Although SEER-MoE also selectively activates critical experts to alleviate computational burden, its absence of aggregation designs leads to severe interference in expert selection, substantially degrading convergence speed.
PFL-MoE and FedMoE, on the other hand, enable client-specific expert selection and aggregates shared experts, but still overlooks the routing divergence among clients, which leads to inconsistent updates of gating parameters for expert selection in each client, resulting in unstable convergence. 
In contrast, sparsity-aware aggregation in \name explicitly accounts for discrepancies in client-specific routing preferences and expert selection, thereby achieving faster convergence across heterogeneous settings.

\begin{figure}[t]
\centering
\subfloat[AGNews \label{fig:computing_agnews}]{
\includegraphics[width=0.49\linewidth]{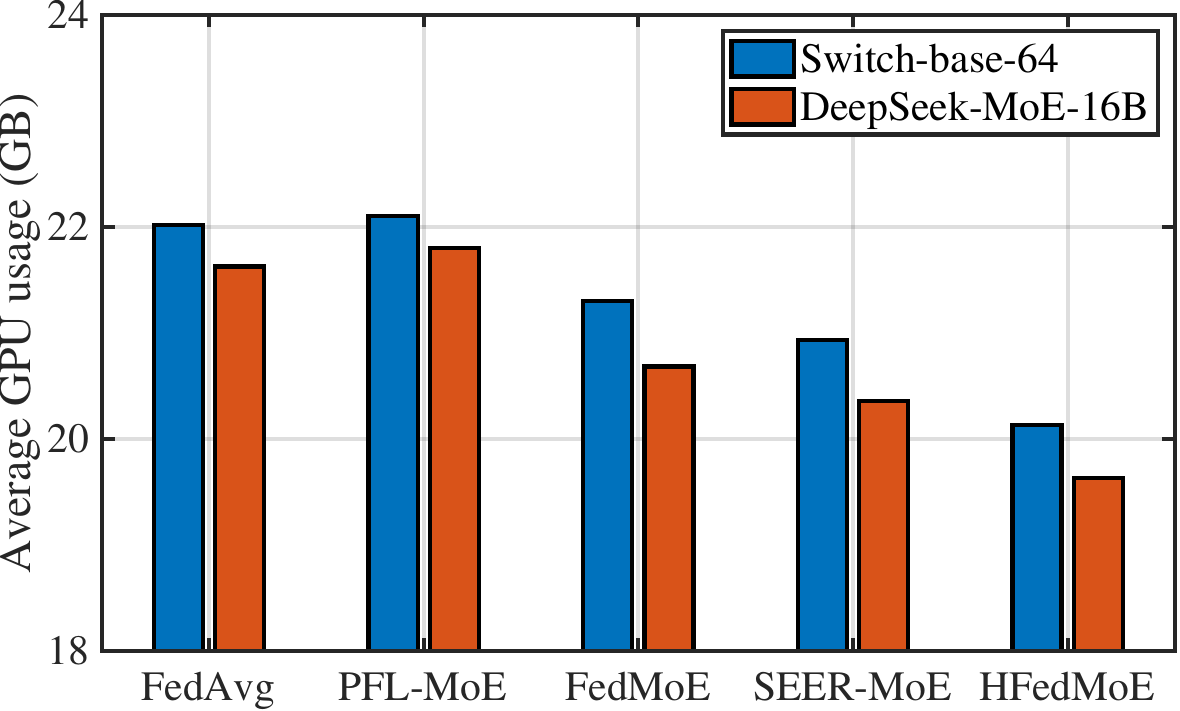}
}
\subfloat[MMLU \label{fig:computing_mmlu}]{
\includegraphics[width=0.49\linewidth]{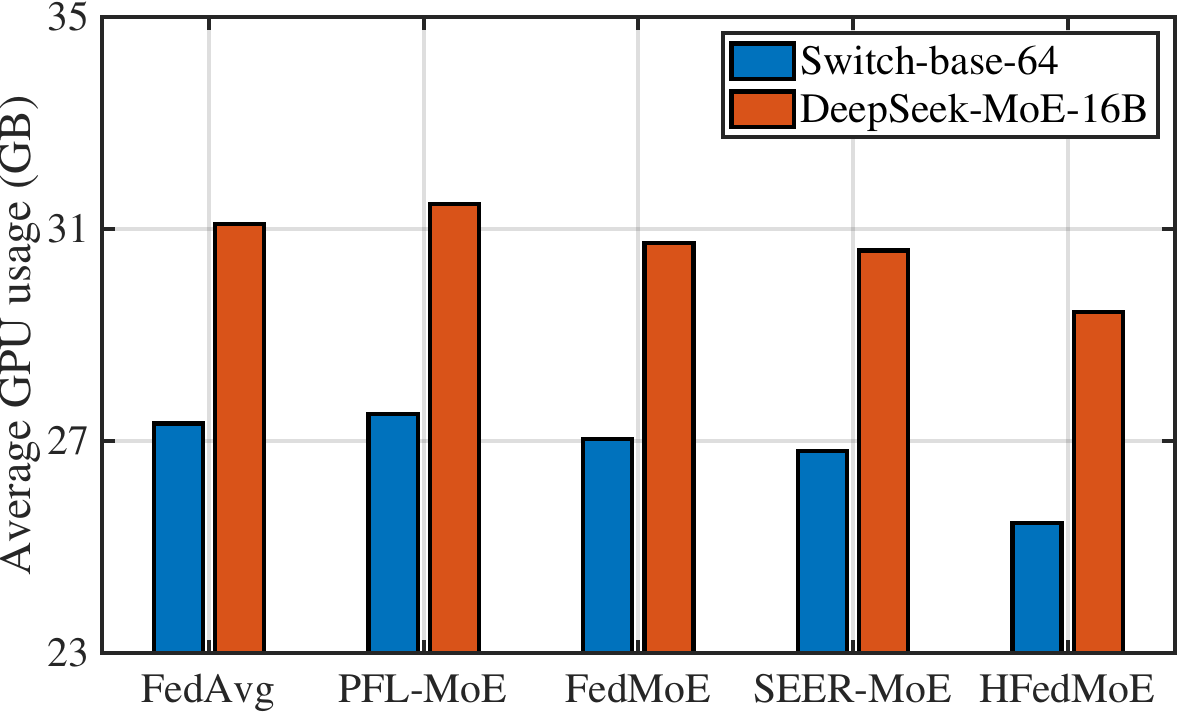}
}
\vspace{-1ex}
\caption{The average GPU usage on the AGNews and MMLU dataset under computing heterogeneity.}
\label{fig:computing_efficiency}
\vspace{-2ex}
\end{figure}

\subsubsection{\textbf{\rev{The Computing Efficiency}}}
Fig.~\ref{fig:computing_efficiency} compares the \waitrev{the average GPU usage} of \name and other four benchmarks on AGNews and MMLU datasets when fine-tuning Switch-base-64 and DeepSeek-MoE-16B models \needrev{under heterogeneous computing constraints}. 
As shown in the figure, \name demonstrates superior computational efficiency: beyond the inherent efficiency of sparse expert activation, it further reduces average GPU usage by \needrev{10\%} compared to the FedAvg and PFL-MoE baselines without computing constraints. 
This advantage stems from \name's expert importance identification that quantifies experts' contribution to local performance, ensuring the limited computational resources are allocated to the most critical experts, especially under limited computing budgets.
By pruning redundant computations from low-impact experts, \name’s resource-aware expert selection strategy dynamically adjusts active expert subsets under local computing constraints, allowing clients with heterogeneous resources to participate effectively participate while preserving representative expert diversity.
In contrast, while FedMoE and SEER-MoE significantly reduce client's computing burden after the selection of a compact expert subset, acquiring stable expert subsets remains computationally challenged, leading to noticeably higher average GPU usage.

\begin{figure}[t]
\centering
\subfloat[AGNews \label{fig:computing_limit_agnews}]{
\includegraphics[width=0.49\linewidth]{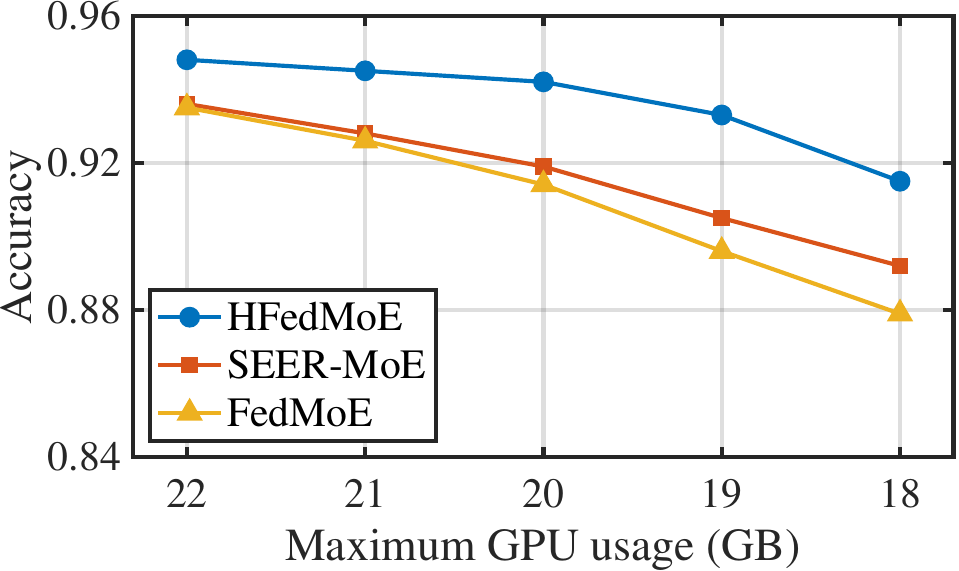}
}
\subfloat[MMLU \label{fig:computing_limit_mmlu_st}]{
\includegraphics[width=0.49\linewidth]{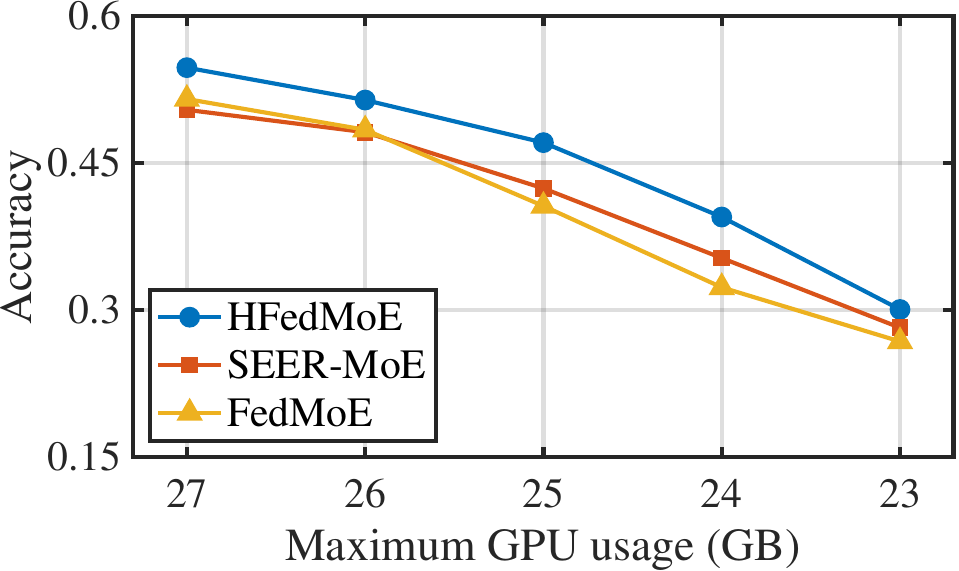}
} \\
\vspace{-2ex}
\subfloat[AGNews \label{fig:computing_limit_agnews_dp}]{
\includegraphics[width=0.49\linewidth]{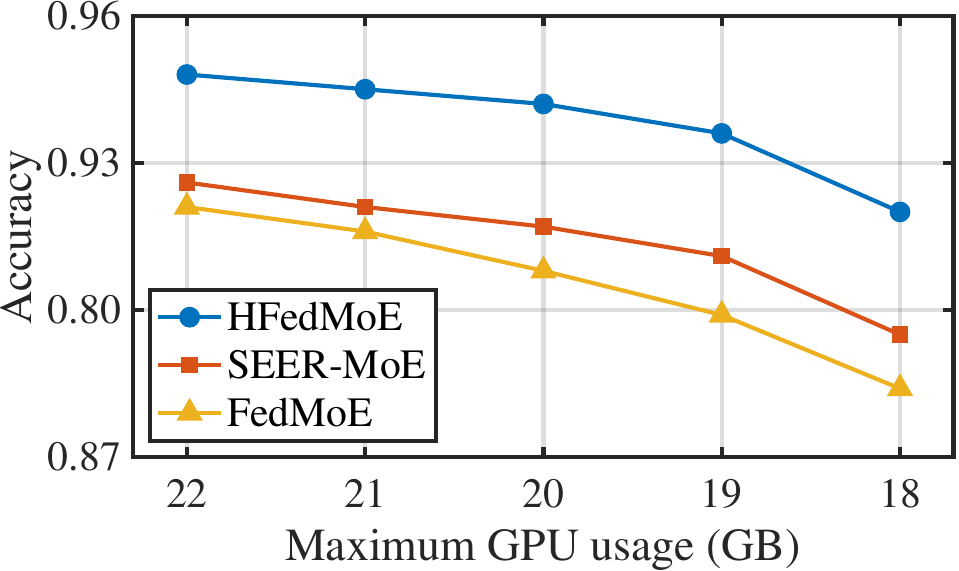}
}
\subfloat[MMLU \label{fig:computing_limit_mmlu}]{
\includegraphics[width=0.49\linewidth]{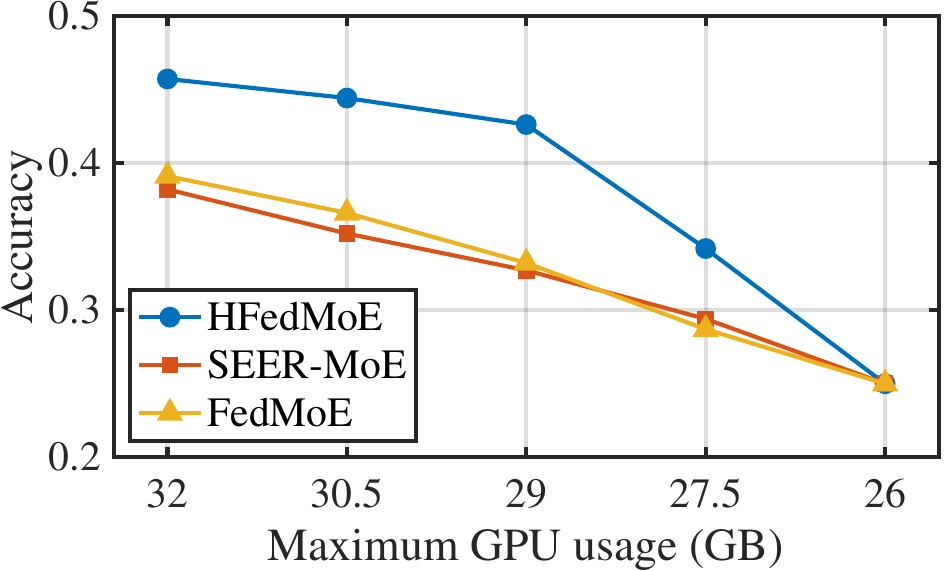}
}
\vspace{-1ex}
\caption{The performance degradation under computing constraints for Switch-base-64 (Fig.~\ref{fig:computing_limit}(a)-(b)) and DeepSeek-MoE-16B (Fig.~\ref{fig:computing_limit}(c)-(d)) models. The computing limitations are imposed by limiting the maximum GPU usage for each client with heterogeneous computing resources.}
\label{fig:computing_limit}
\vspace{-1ex}
\end{figure}

\subsubsection{\textbf{The Impact of Computational Limitation}}
Fig.~\ref{fig:computing_limit} reports the impact of computing limitation for \name on converged test accuracy using AGNews and MMLU tasks.
While all baseline methods exhibit inevitable accuracy degradation under reduced computing capacity, \name demonstrates remarkable robustness by maintaining over \needrev{95\%} of its full-performance accuracy, even when half of clients face substantial resource constraints. This is primarily attributed to \name's resource-aware expert selection strategy, which effectively allocates limited resources to the most informative local updates, and its sparsity-aware aggregation for experts and gating networks, which explicitly handles partial expert updates and divergent routing preferences.
Although FedMoE and SEER-MoE also incorporate expert importance estimation to reduce memory, its coarse-grained importance estimation fails to account for input-specific contributions across samples while concurrent expert activations across samples during preliminary training severely increases the client burden, leading to severe accuracy degradation under tight computing budgets.


\begin{table*}[t]
\centering
\scalebox{0.9}{
  \begin{tabular}{c||cc|cc||cc|cc}
    \hline
    \multirow{3}{*}{$\lambda$} & \multicolumn{4}{c||}{\textbf{AGNews}} & \multicolumn{4}{c}{\textbf{MMLU}} \\
    \cline{2-9}
    & \multicolumn{2}{c|}{Switch-base-64} & \multicolumn{2}{c||}{DeepSeek-MoE-16B} & \multicolumn{2}{c|}{Switch-base-64} & \multicolumn{2}{c}{DeepSeek-MoE-16B} \\
    & Convergence & Accuracy & Convergence & Accuracy & Convergence & Accuracy & Convergence & Accuracy \\
    \hline
    \texttt{1.0}  & 21 & 0.9434 & 38 & 0.9391  & 8 & 0.4871 & 11 & 0.3963 \\
    \texttt{0.9}  & 23 & 0.9453 & 36 & 0.9409  & \textbf{7} & \textbf{0.5132} & \textbf{13} & \textbf{0.4421} \\
    \texttt{0.8}  & \textbf{22} & \textbf{0.9460} & \textbf{35} & \textbf{0.9419}  & 11 & 0.5001 & 16 & 0.4290 \\
    \texttt{0.7}  & 24 & 0.9445 & 36 & 0.9401  & 12 & 0.4861 & 20 & 0.4421 \\
    \texttt{0.6}  & 26 & 0.9416 & 39 & 0.9412  & 16 & 0.5097 & 18 & 0.4355 \\
    \hline
  \end{tabular}
}
\caption{The performance of \name with different weighted combination of expert importance.}
\label{fig:hyperparameter_lambda}
\end{table*}


\begin{table*}[t]
\centering
\scalebox{0.9}{
  \begin{tabular}{c||cc|cc||cc|cc}
    \hline
    \multirow{3}{*}{$\tau$} & \multicolumn{4}{c||}{\textbf{AGNews}} & \multicolumn{4}{c}{\textbf{MMLU}} \\
    \cline{2-9}
    & \multicolumn{2}{c|}{Switch-base-64} & \multicolumn{2}{c||}{DeepSeek-MoE-16B} & \multicolumn{2}{c|}{Switch-base-64} & \multicolumn{2}{c}{DeepSeek-MoE-16B} \\
    & Convergence & Accuracy & Convergence & Accuracy & Convergence & Accuracy & Convergence & Accuracy \\
    \hline
    \texttt{0.0}  & 20 & 0.9417 & 37 & 0.9401 & 9 & 0.5001 & 22 & 0.4421 \\
    \texttt{0.05} & \textbf{22} & \textbf{0.9460} & \textbf{35} & \textbf{0.9419} & \textbf{7} & \textbf{0.5132} & \textbf{13} & \textbf{0.4421} \\
    \texttt{0.1}  & 24 & 0.9437 & 40 & 0.9385 & 8 & 0.4871 & 17 & 0.4291 \\
    \texttt{0.15} & 24 & 0.9408 & 42 & 0.9397 & 13 & 0.4936 & 20 & 0.4421 \\
    \texttt{0.2}  & 27 & 0.9395 & 49 & 0.9364 & 19 & 0.4796 & 8 & 0.3137 \\
    \hline
  \end{tabular}
}
\caption{The performance of \name with different usage threshold for actively trained experts.}
\vspace{-2ex}
\label{fig:hyperparameter_tau}
\end{table*}

\vspace{-1ex}
\subsection{The Optimal Hyper-parameters}
\subsubsection{\textbf{Varying Weighted Combination of Expert Importance}}
Table~\ref{fig:hyperparameter_lambda} demonstrates the impact of varying weighting coefficient $\lambda$ in \name, which controls the balance between cumulative and specific importance in expert importance identification. Experiments on the AGNews and MMLU datasets using Switch-base-64 and DeepSeek-MoE-16B models demonstrate that convergence time and test accuracy exhibit a clear dependency on the choice of $\lambda$.
When $\lambda$ is small, expert selection relies more on specific importance, emphasizing adaptation to specific and often hard-to-classify samples. This leads to faster initial convergence as critical responses are preserve during fine-tuning, but also increases the risk of overfitting to suboptimal routing patterns, resulting in lower accuracy.
In contrast, a larger $\lambda$ increases the weight of cumulative importance, encouraging effective representation across samples and improving generalization, though at the cost of slower convergence due to reduced expert diversity.
Moreover, we observe that the optimal value of $\lambda$ varies across datasets: a smaller value ($\lambda=0.8$) performs better on AGNews, while a larger setting ($\lambda=0.9$) achieves the best trade-off between convergence and accuracy on MMLU, \rev{suggesting that datasets with higher heterogeneity benefit from higher cumulative importance to stabilize expert selection.}

\subsubsection{\textbf{Varying Usage Threshold for Actively Trained Experts}}
Table~\ref{fig:hyperparameter_tau} presents the impact of varying the usage threshold $\tau$ on convergence time and test accuracy when aggregating Switch-base-64 and DeepSeek-MoE-16B models on the AGNews and MMLU datasets.
The results highlight the advantage of excluding undertrained experts from the global aggregation. 
Without selective expert aggregation ($\tau=0$), partially trained or entirely untouched experts are still aggregated indiscriminately, introducing interference among inconsistently optimized experts and degrading the generalization of the global model. Such aggregation hinders the alignment of expert specialization across clients, leading to slower convergence and reduced generalization capability of the global model.
In contrast, setting $\tau$ too high restricts aggregation to only a small subset of experts, limiting global knowledge sharing and diminishing the diversity of learned representations, thus resulting in suboptimal accuracy.
\rev{A moderate threshold ($\tau=0.05$)} provides the best balance by filtering out inactively trained experts while retaining diverse and actively trained experts for aggregation, leading to more stable convergence and highest accuracy on AGNews and MMLU. 

\vspace{-1ex}
\subsection{Ablation Experiments}
\subsubsection{\textbf{Expert Importance Identification}}
\begin{figure}[t]
\centering
\vspace{-2ex}
\subfloat[AGNews]{
\includegraphics[width=0.49\linewidth]{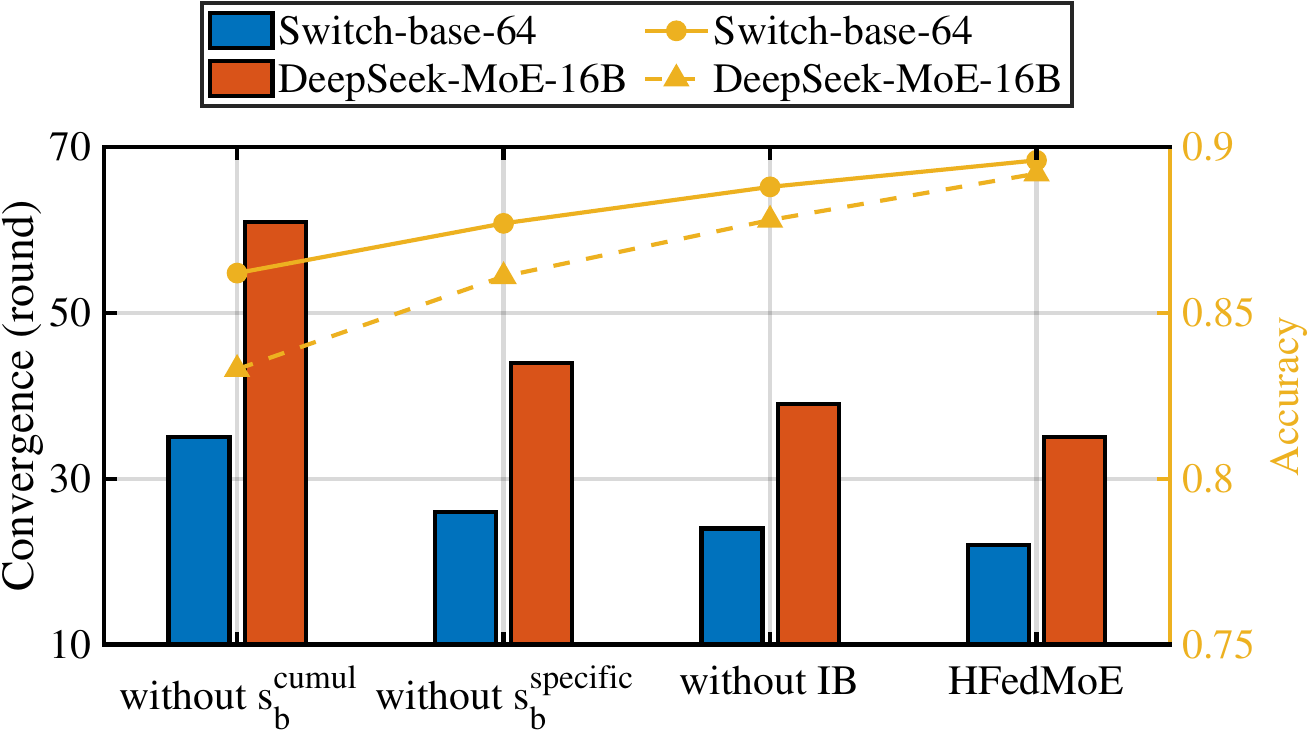}
}
\subfloat[MMLU]{
\includegraphics[width=0.49\linewidth]{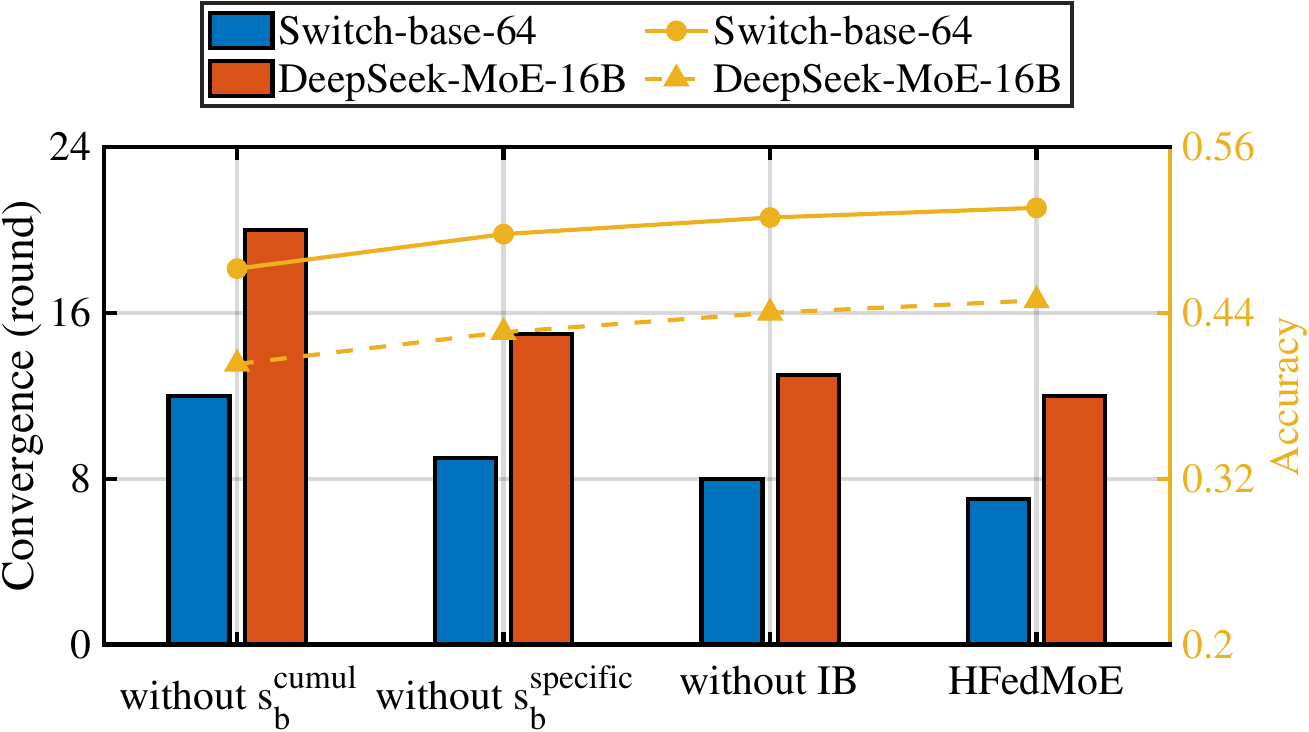}
}
\vspace{-1ex}
\caption{Expert Importance Identification on AGNews and MMLU dataset.}
\label{fig:ablation_importance}
\vspace{-2ex}
\end{figure}

Fig.~\ref{fig:ablation_importance} illustrates the impact of expert importance estimation on convergence time and test accuracy for fine-tuning Switch-base-64 and DeepSeek-MoE-16B models. 
The results demonstrate that both cumulative and specific importance of experts contribute notably to performance improvements, while the absence of the IB optimization objective yields less discriminative expert selection and reduced accuracy.
Specifically, cumulative importance captures the overlap of expert usage across input samples, enables substantial computation sharing and redundant computation reduction, thereby accelerating convergence x1.35 under constrained computing resources. Incorporating specific importance further captures each expert’s distinct contribution to local data, allowing enhanced adaptation to better aligns with client-specific data distributions. This yields an additional \needrev{x1.2} speedup and improves test accuracy by approximately \needrev{2\%},  
showcasing the superior performance of the proposed expert importance metrics.

\begin{figure}[t]
\centering
\vspace{-1ex}
\subfloat[Experts in Switch-base-64]{
\includegraphics[width=0.49\linewidth]{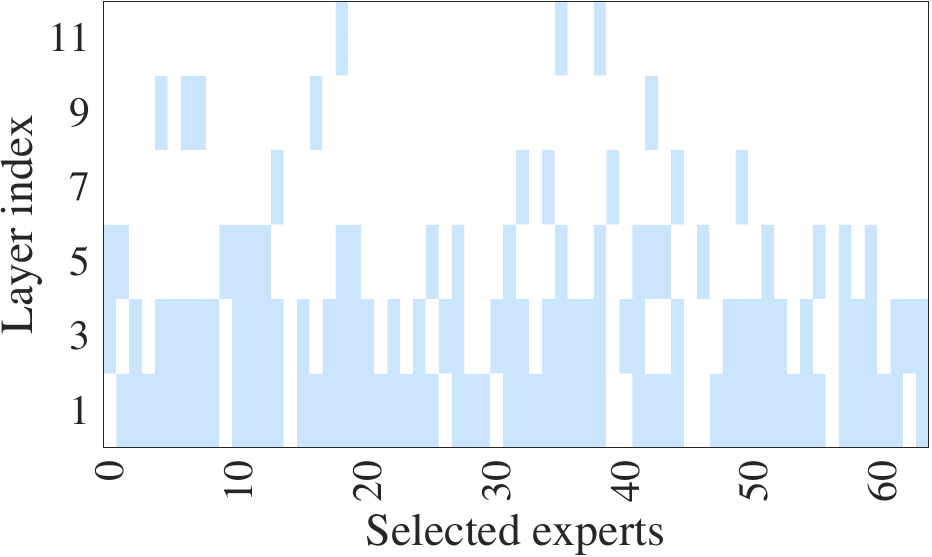}
}
\subfloat[Experts in DeepSeek-MoE]{
\includegraphics[width=0.49\linewidth]{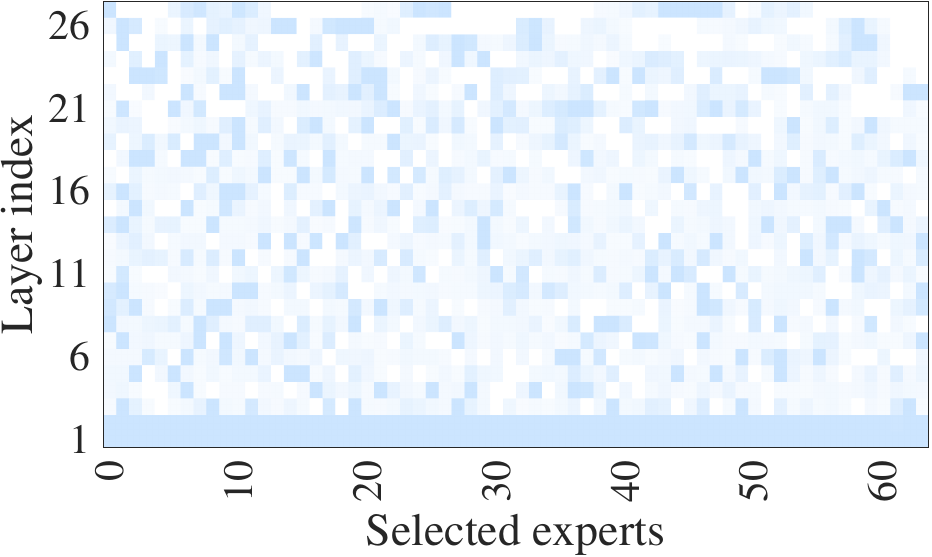}
} \\
\vspace{-2ex}
\subfloat[Experts in Switch-base-64 \label{fig:ablation_selection_switch_mmlu}]{
\includegraphics[width=0.49\linewidth]{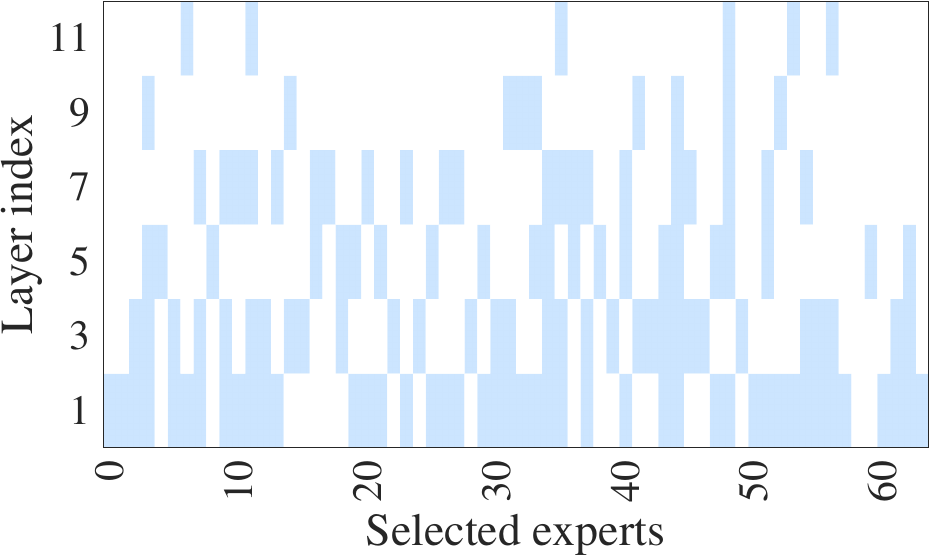}
}
\subfloat[Experts in DeepSeek-MoE \label{fig:ablation_selection_deepseek_mmlu}]{
\includegraphics[width=0.49\linewidth]{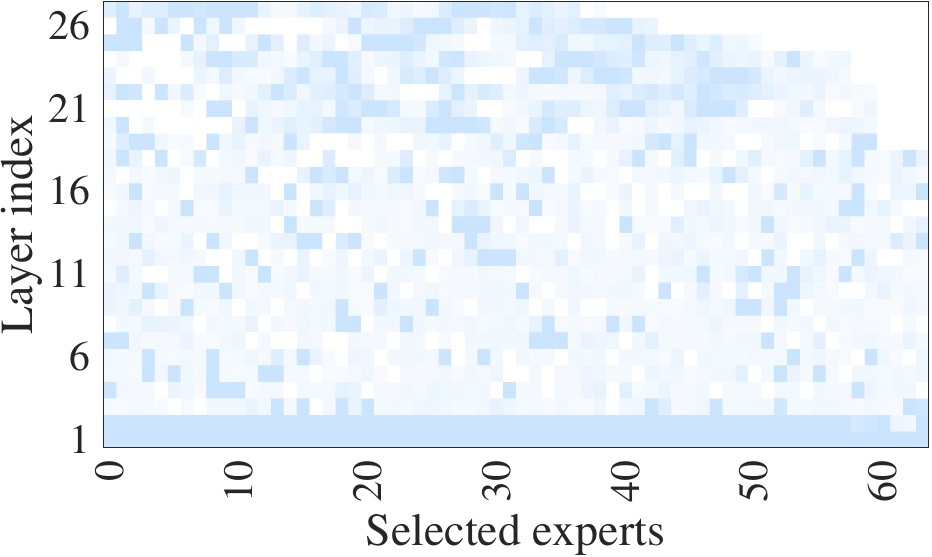}
}
\vspace{-1ex}
\caption{Resource-aware Expert Selection on \rev{AGNews (Fig.~\ref{fig:ablation_selection}(a)-(b)) and MMLU (Fig.~\ref{fig:ablation_selection}(c)-(d))} datasets using Switch-base-64 and DeepSeek-MoE-16B models.}
\label{fig:ablation_selection}
\vspace{-2ex}
\end{figure}

\subsubsection{\textbf{Resource-aware Expert Selection}}
Fig.~\ref{fig:ablation_selection} shows the number and distribution of experts activated during fine-tuning under limited computing resources. 
\name activates about \needrev{25\% and 78\%} of the experts during fine-tuning on the Switch-base-64 and DeepSeek-MoE-16B models, respectively, while maintaining over 95\% of their original test accuracy. These results underscore the capability of \name to achieve efficient resource utilization with marginal performance degradation, highlighting its potential for deployment on resource-constrained devices.
This efficiency is driven by the proposed resource-aware expert selection, which prioritizes high-impact experts while reducing redundancy on experts with minimal impact on performance.
Moreover, it is observed that expert selection is not uniformly applied across layers: deeper layers tend to retain fewer experts, whereas more functionally critical layers maintain a higher degree of activation. 
This adaptive allocation maximizes computational efficiency and aligns with prior observations~\cite{men2024shortgpt} suggesting that deeper layers in large-scale models are often more redundant.

\subsubsection{\textbf{Sparsity-aware Model Aggregation}}
\begin{figure}[t]
\vspace{-1ex}
\centering
\subfloat[AGNews]{
\includegraphics[width=0.49\linewidth]{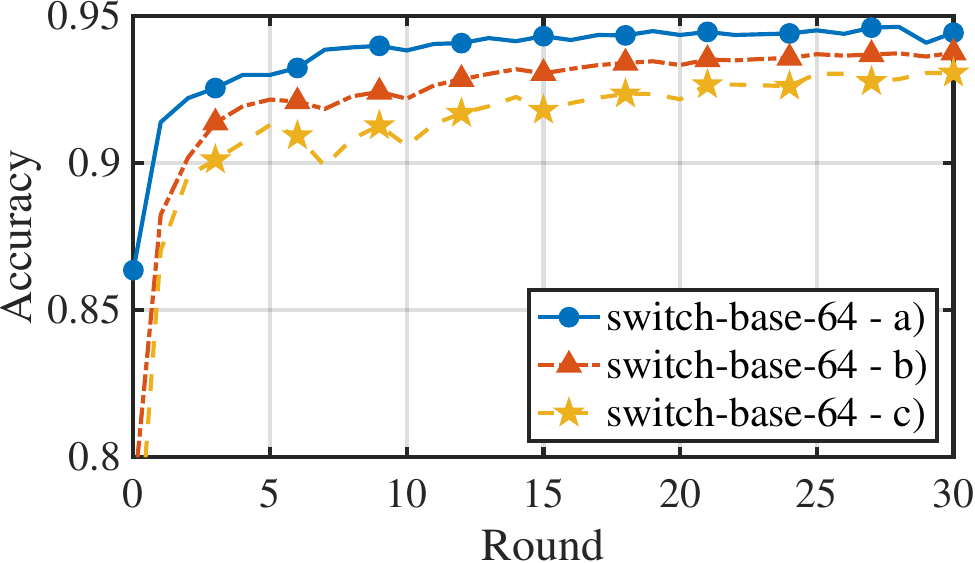}
}
\subfloat[MMLU]{
\includegraphics[width=0.49\linewidth]{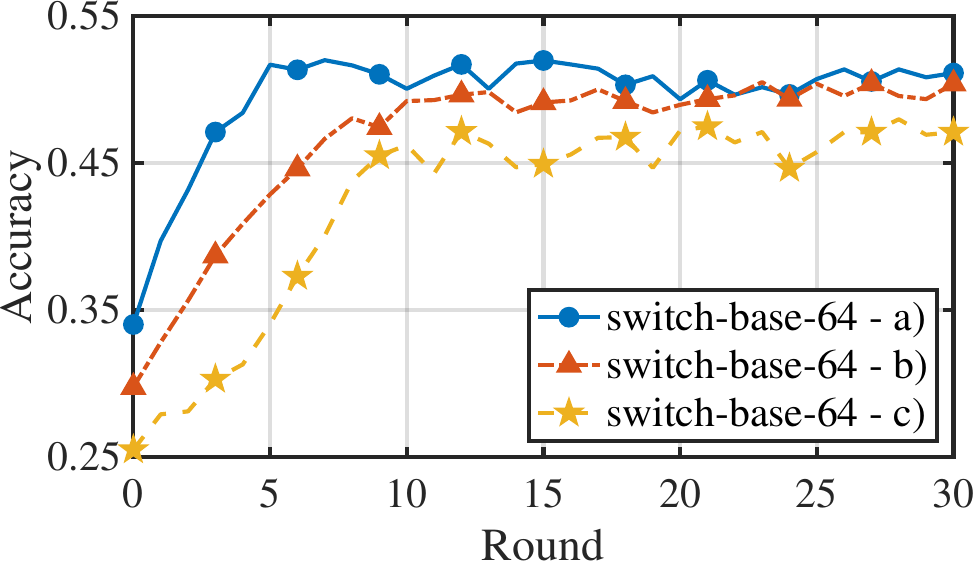}
} \\
\vspace{-1ex}
\subfloat[AGNews]{
\includegraphics[width=0.49\linewidth]{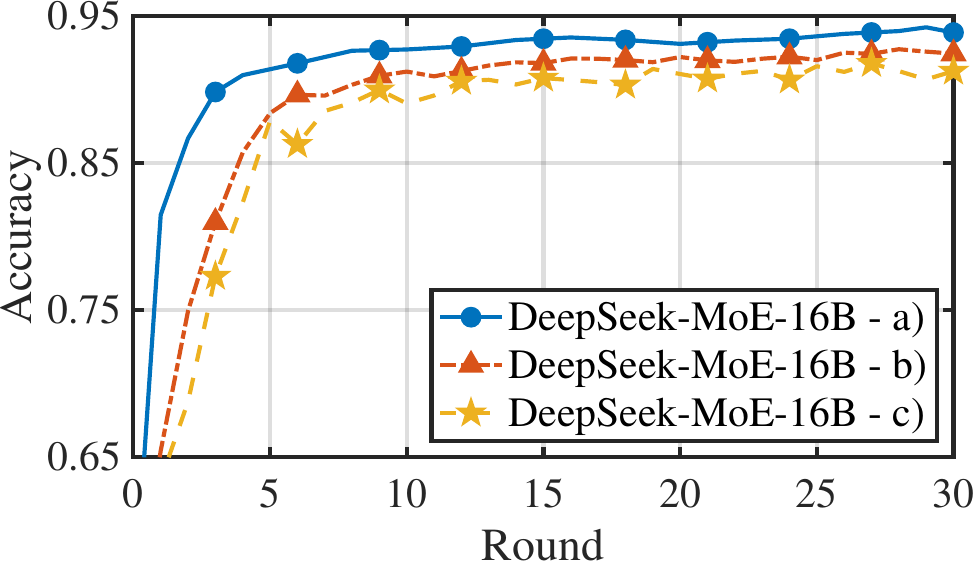}
}
\subfloat[MMLU]{
\includegraphics[width=0.49\linewidth]{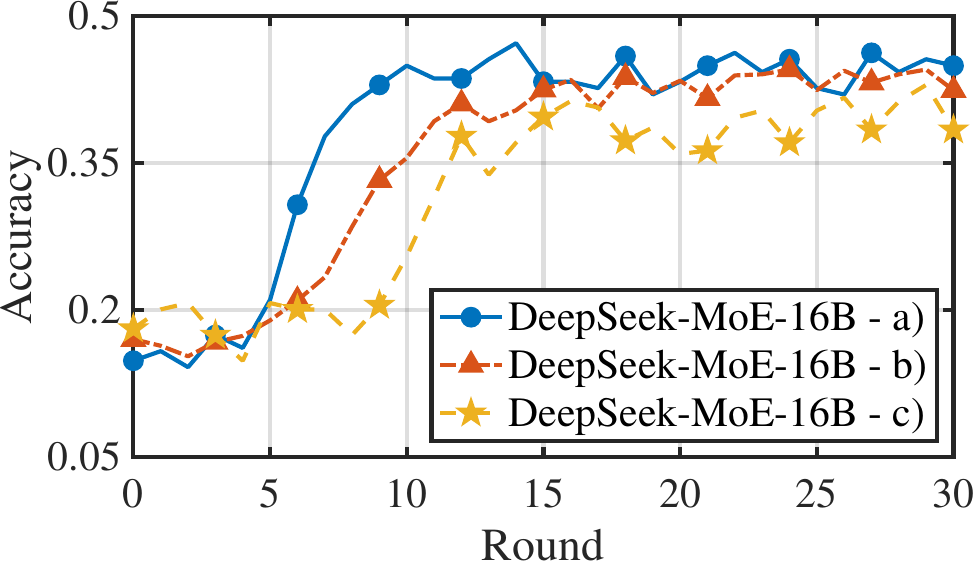}
}
\vspace{-1ex}
\caption{Sparsity-aware Model Aggregation on AGNews and \needrev{MMLU} dataset using Switch-base-64 (Fig.~\ref{fig:ablation_aggregation}(a)-(b)) and DeepSeek-MoE-16B (Fig.~\ref{fig:ablation_aggregation}(c)-(d)) models.}
\label{fig:ablation_aggregation}
\vspace{-2ex}
\end{figure}

Fig.~\ref{fig:ablation_aggregation} presents the fine-tuning performance of model aggregation on the Switch-base-64 and DeepSeek-MoE-16B models under heterogeneous computing resources. Three aggregation conditions are compared: a). selectively aggregate experts while aggregating gating parameters with importance-weighted contributions, b). selective expert aggregation without accounting for divergent routing preferences across clients, and c). standard FedAvg aggregation, which aggregates all expert and gating parameters uniformly across clients.
Compared to the standard aggregation that assumes uniform model structures and meaningful updates for all parameter, \name achieves consistently better accuracy and faster convergence on both Switch-base-64 and DeepSeek-MoE-16B. This improvement stems from the proposed sparsity-aware model aggregation, which filters out unreliable expert updates and aggregates only active experts, effectively mitigating the interference caused by partial expert updates across clients. 
Moreover, importance-weighted gating aggregation alleviates the inconsistent routing behaviors by adjust each client’s contribution to the global gating network according to its expert preferences and routing consistency, yielding additional benefits on test accuracy and convergence.

\section{Related Work} \label{sec:related_work}

{\bf{MoE-based LLM in Federated Learning: }}
By selectively activating a subset of experts, MoE provides a promising solution to reduce computational overhead while preserving model's representational capacity.
Leveraging this efficiency, MoE has been adopted as a foundational framework in the development of LLMs~\cite{dai2024deepseekmoe, fedus2022switch, jiang2024mixtral}, achieving superior performance at low computational costs.
Further investigations have extended MoE to personalized FL~\cite{guo2021pfl, dun2023fedjets, feng2025pm}, where client-specific expert activation allows each client to fine-tune only the experts most relevant to its local data.
While improving local adaptation, these approaches often rely on shared gating networks for expert selection, overlooking each expert’s contribution to critical local features and resulting in suboptimal performance.
Moreover, most existing works fail to account for the limited computing resources in client devices, significantly limiting the deployment of MoE-based LLMs in heterogeneous federated environments.

{\bf{MoE-based LLM Fine-tuning for Computational Reduction: }}
To further reduce computing overhead of MoE-based LLMs, recent work has investigated dedicated expert selection strategies.
Current resource-constrained FL methods, such as FedMoE~\cite{mei2024fedmoe}, use preliminary training to identify client-specific expert subsets, thereby alleviating the client's memory pressure during subsequent fine-tuning. However, the preliminary training phase to acquire stable expert subsets remains computationally demanding due to the elevated computation imposed by concurrent expert activations across diverse samples, thus limiting its ability to mitigate computing resource constraints. Moreover, failing to handle routing divergence among clients and accurately estimate expert importance restrict the attainment of reliable and optimal structures necessary for effective resource mitigation.
Alternatively, pruning-based approaches~\cite{muzio2024seer, jin2025moe, he2024towards, lu2024not} reduce resource consumption during fine-tuning by eliminating experts with low average activation frequency.
\peerrev{However, they similarly fail to account for the concurrent activation in expert selection across diverse inputs and fail to accommodate the heterogeneous computing capacities across client devices, hindering their ability to achieve an optimal trade-off between performance and efficiency in federated settings, especially under client-specific resource constraints during fine-tuning.}

{\bf{Model Aggregation for Federated MoE Fine-tuning: }}
Despite growing interest in integrating MoE models into FL, effectively aggregating structurally heterogeneous local models while preserving global performance poses significant challenges. 
Current FL approaches~\cite{mcmahan17a, lin2024efficient, zhang2024towards} usually assume that all clients share identical model structures, whereas distinct expert subsets and divergent routing preferences render naive parameter aggregation ineffective.
Several recent studies~\cite{guo2021pfl, dun2023fedjets, jin2025moe} attempt to alleviate structural heterogeneity among experts but still overlook diverse expert activation patterns during aggregation, thus failing to handle partial expert updates and inconsistent gating networks and leading to degraded performance in MoE-based federated learning.
\rev{Although some approaches~\cite{zhan2024fedmoe, hu2025fft, yi2024pfedmoe} preserve local adaptability by maintaining private gating networks and aggregating only expert parameters, the resulting partial expert updates across clients lead to misaligned optimization, thus weakening expert specialization and impairing model generalization.}

\section{Conclusion}\label{sec:conclusion}
In this paper, we have proposed a heterogeneous federated learning framework, named \name, to enhance the effectiveness of MoE-based LLM fine-tuning across heterogeneous clients with diverse computing capabilities.
First, the expert importance identification scheme quantifies each expert’s contribution to the fine-tuning performance across diverse clients. Second, based on each expert's importance, the resource-aware expert selection method ensures that the critical experts are dynamically chosen under each client’s computing budget.
Finally, the selective expert aggregation strategy further enhances model generalization, mitigating the impact of expert structural heterogeneity.
Extensive experimental results have demonstrated that \name outperforms the state-of-the-art benchmarks, achieving significantly improved convergence with minimal accuracy degradation. 


\ifCLASSOPTIONcaptionsoff
  \newpage
\fi



%



\bibliographystyle{IEEEtran}
\bibliography{reference}

\begin{thebibliography}{10}
\providecommand{\url}[1]{#1}
\csname url@samestyle\endcsname
\providecommand{\newblock}{\relax}
\providecommand{\bibinfo}[2]{#2}
\providecommand{\BIBentrySTDinterwordspacing}{\spaceskip=0pt\relax}
\providecommand{\BIBentryALTinterwordstretchfactor}{4}
\providecommand{\BIBentryALTinterwordspacing}{\spaceskip=\fontdimen2\font plus
\BIBentryALTinterwordstretchfactor\fontdimen3\font minus \fontdimen4\font\relax}
\providecommand{\BIBforeignlanguage}[2]{{%
\expandafter\ifx\csname l@#1\endcsname\relax
\typeout{** WARNING: IEEEtran.bst: No hyphenation pattern has been}%
\typeout{** loaded for the language `#1'. Using the pattern for}%
\typeout{** the default language instead.}%
\else
\language=\csname l@#1\endcsname
\fi
#2}}
\providecommand{\BIBdecl}{\relax}
\BIBdecl

\bibitem{achiam2023gpt}
J.~Achiam, S.~Adler, S.~Agarwal, L.~Ahmad, I.~Akkaya, F.~L. Aleman, D.~Almeida, J.~Altenschmidt, S.~Altman, S.~Anadkat \emph{et~al.}, ``{Gpt-4 Technical Report},'' \emph{arXiv preprint arXiv:2303.08774}, Mar. 2023.

\bibitem{lin2023pushing}
Z.~Lin, G.~Qu, Q.~Chen, X.~Chen, Z.~Chen, and K.~Huang, ``{Pushing Large Language Models to the 6G Edge: Vision, Challenges, and Opportunities},'' \emph{IEEE Communication Magazine}, 2023.

\bibitem{duan2025leed}
T.~Duan, Z.~Zhang, S.~Guo, D.~Huang, Y.~Zhao, Z.~Lin, Z.~Fang, D.~Luan, H.~Cui, and Y.~Cui, ``Leed: A highly efficient and scalable llm-empowered expert demonstrations framework for multi-agent reinforcement learning,'' \emph{arXiv preprint arXiv:2509.14680}, 2025.

\bibitem{touvron2023llama}
H.~Touvron, T.~Lavril, G.~Izacard, X.~Martinet, M.-A. Lachaux, T.~Lacroix, B.~Rozi{\`e}re, N.~Goyal, E.~Hambro, F.~Azhar \emph{et~al.}, ``{Llama: Open and Efficient Foundation Language Models},'' \emph{arXiv preprint arXiv:2302.13971}, Feb. 2023.

\bibitem{lin2024splitlora}
Z.~Lin, X.~Hu, Y.~Zhang, Z.~Chen, Z.~Fang, X.~Chen, A.~Li, P.~Vepakomma, and Y.~Gao, ``{SplitLoRA: A Split Parameter-Efficient Fine-Tuning Framework for Large Language Models},'' \emph{arXiv preprint arXiv:2407.00952}, 2024.

\bibitem{liu2024deepseek}
A.~Liu, B.~Feng, B.~Xue, B.~Wang, B.~Wu, C.~Lu, C.~Zhao, C.~Deng, C.~Zhang, C.~Ruan \emph{et~al.}, ``{Deepseek-v3 Technical Report},'' \emph{arXiv preprint arXiv:2412.19437}, Feb. 2025.

\bibitem{fang2024automated}
Z.~Fang, Z.~Lin, Z.~Chen, X.~Chen, Y.~Gao, and Y.~Fang, ``{Automated Federated Pipeline for Parameter-Efficient Fine-Tuning of Large Language Models},'' \emph{{IEEE} Trans. Mobile Comput.}, 2026.

\bibitem{lin2025hsplitlora}
Z.~Lin, Y.~Zhang, Z.~Chen, Z.~Fang, X.~Chen, P.~Vepakomma, W.~Ni, J.~Luo, and Y.~Gao, ``{HSplitLoRA: A Heterogeneous Split Parameter-Efficient Fine-Tuning Framework for Large Language Models},'' \emph{arXiv preprint arXiv:2505.02795}, 2025.

\bibitem{duan2025llm}
T.~Duan, Z.~Zhang, Z.~Lin, S.~Guo, X.~Guan, G.~Wu, Z.~Fang, H.~Meng, X.~Du, J.-Z. Zhou \emph{et~al.}, ``{LLM-Driven Stationarity-Aware Expert Demonstrations for Multi-Agent Reinforcement Learning in Mobile Systems},'' \emph{arXiv preprint arXiv:2511.19368}, 2025.

\bibitem{fang2025dynamic}
Z.~Fang, Z.~Lin, S.~Hu, Y.~Tao, Y.~Deng, X.~Chen, and Y.~Fang, ``Dynamic uncertainty-aware multimodal fusion for outdoor health monitoring,'' \emph{arXiv preprint arXiv:2508.09085}, 2025.

\bibitem{karargyris2023federated}
A.~Karargyris, R.~Umeton, M.~J. Sheller, A.~Aristizabal, J.~George, A.~Wuest, S.~Pati, H.~Kassem, M.~Zenk, U.~Baid \emph{et~al.}, ``{Federated Benchmarking of Medical Artificial Intelligence with MedPerf},'' \emph{Nature Machine Intelligence}, vol.~5, no.~7, pp. 799--810, Jul. 2023.

\bibitem{mcmahan17a}
B.~McMahan, E.~Moore, D.~Ramage, S.~Hampson, and B.~A.~y. Arcas, ``{Communication-Efficient Learning of Deep Networks from Decentralized Data},'' in \emph{Proc. AISTATS}, vol.~54.\hskip 1em plus 0.5em minus 0.4em\relax PMLR, Apr. 2017, pp. 1273--1282.

\bibitem{luo2023optimization}
B.~Luo, P.~Han, P.~Sun, X.~Ouyang, J.~Huang, and N.~Ding, ``{Optimization Design for Federated Learning in Heterogeneous 6G Networks},'' \emph{{IEEE} Netw.}, vol.~37, no.~2, pp. 38--43, Apr. 2023.

\bibitem{hu2024accelerating}
M.~Hu, J.~Zhang, X.~Wang, S.~Liu, and Z.~Lin, ``{Accelerating Federated Learning with Model Segmentation for Edge Networks},'' \emph{{IEEE} Trans. Green Commun. Netw.}, 2024.

\bibitem{zhang2024satfed}
Y.~Zhang, Z.~Lin, Z.~Chen, Z.~Fang, W.~Zhu, X.~Chen, J.~Zhao, and Y.~Gao, ``Satfed: A resource-efficient leo satellite-assisted heterogeneous federated learning framework,'' \emph{Engineering}, 2024.

\bibitem{bonawitz2019towards}
K.~Bonawitz, H.~Eichner, W.~Grieskamp, D.~Huba, A.~Ingerman, V.~Ivanov, C.~Kiddon, J.~Kone{\v{c}}n{\`y}, S.~Mazzocchi, B.~McMahan \emph{et~al.}, ``{Towards Federated Learning at Scale: System Design},'' \emph{Proc. SysML}, vol.~1, pp. 374--388, Mar. 2019.

\bibitem{lin2024fedsn}
Z.~Lin, Z.~Chen, Z.~Fang, X.~Chen, X.~Wang, and Y.~Gao, ``Fedsn: A federated learning framework over heterogeneous leo satellite networks,'' \emph{{IEEE} Trans. Mobile Comput.}, 2024.

\bibitem{zhang2025lcfed}
Y.~Zhang, H.~Chen, Z.~Lin, Z.~Chen, and J.~Zhao, ``{LCFed: An Efficient Clustered Federated Learning Framework for Heterogeneous Data},'' \emph{arXiv preprint arXiv:2501.01850}, 2025.

\bibitem{wang2024flora}
Z.~Wang, Z.~Shen, Y.~He, G.~Sun, H.~Wang, L.~Lyu, and A.~Li, ``{Flora: Federated Fine-tuning Large Language Models with Heterogeneous Low-Rank Adaptations},'' \emph{Proc. NeurIPS}, vol.~37, pp. 22\,513--22\,533, Dec. 2024.

\bibitem{lin2024adaptsfl}
Z.~Lin, G.~Qu, W.~Wei, X.~Chen, and K.~K. Leung, ``{Adaptsfl: Adaptive Split Federated Learning in Resource-Constrained Edge Networks},'' \emph{{IEEE} Trans. Netw.}, 2024.

\bibitem{jiang2023computation}
Z.~Jiang, Y.~Xu, H.~Xu, Z.~Wang, J.~Liu, Q.~Chen, and C.~Qiao, ``{Computation and Communication Efficient Federated Learning with Adaptive Model Pruning},'' \emph{{IEEE} Trans. Mobile Comput.}, vol.~23, no.~3, pp. 2003--2021, Mar. 2023.

\bibitem{lin2024efficient}
Z.~Lin, G.~Zhu, Y.~Deng, X.~Chen, Y.~Gao, K.~Huang, and Y.~Fang, ``{Efficient Parallel Split Learning over Resource-Constrained Wireless Edge Networks},'' \emph{{IEEE} Trans. Mobile Comput.}, vol.~23, no.~10, pp. 9224--9239, 2024.

\bibitem{lin2025hasfl}
Z.~Lin, Z.~Chen, X.~Chen, W.~Ni, and Y.~Gao, ``{HASFL: Heterogeneity-aware Split Federated Learning over Edge Computing Systems},'' \emph{arXiv preprint arXiv:2506.08426}, 2025.

\bibitem{dhar2024empirical}
N.~Dhar, B.~Deng, D.~Lo, X.~Wu, L.~Zhao, and K.~Suo, ``{An Empirical Analysis and Resource Footprint Study of Ddeploying Large Language Models on Edge Devices},'' in \emph{Proc. ACMSE}, Apr. 2024, pp. 69--76.

\bibitem{xu2025edgellm}
D.~Xu, W.~Yin, H.~Zhang, X.~Jin, Y.~Zhang, S.~Wei, M.~Xu, and X.~Liu, ``{EdgeLLM: Fast On-Device LLM Inference With Speculative Decoding},'' \emph{{IEEE} Trans. on Mobile Comput.}, vol.~24, no.~4, pp. 3256--3273, 2025.

\bibitem{qu2025mobile}
G.~Qu, Q.~Chen, W.~Wei, Z.~Lin, X.~Chen, and K.~Huang, ``{Mobile Edge Intelligence for Large Language Models: A Contemporary Survey},'' \emph{{IEEE} Commun. Surv. Tutor.}, 2025.

\bibitem{dai2024deepseekmoe}
D.~Dai, C.~Deng, C.~Zhao, R.~Xu, H.~Gao, D.~Chen, J.~Li, W.~Zeng, X.~Yu, Y.~Wu \emph{et~al.}, ``{DeepSeekMoE: Towards Ultimate Expert Specialization in Mixture-of-Experts Language Models},'' in \emph{Proc. ACL (Volume 1: Long Papers)}, 2024, pp. 1280--1297.

\bibitem{hu2021lora}
E.~J. Hu, Y.~Shen, P.~Wallis, Z.~Allen-Zhu, Y.~Li, S.~Wang, L.~Wang, and W.~Chen, ``{Lo{RA}: Low-Rank Adaptation of Large Language Models},'' in \emph{Proc. ICLR}, Jan. 2022.

\bibitem{houlsby2019parameter}
N.~Houlsby, A.~Giurgiu, S.~Jastrzebski, B.~Morrone, Q.~De~Laroussilhe, A.~Gesmundo, M.~Attariyan, and S.~Gelly, ``{Parameter-Efficient Transfer Learning for {NLP}},'' in \emph{Proc. ICML}, Jun. 2019, pp. 2790--2799.

\bibitem{lin2025hierarchical}
Z.~Lin, W.~Wei, Z.~Chen, C.-T. Lam, X.~Chen, Y.~Gao, and J.~Luo, ``{Hierarchical Split Federated Learning: Convergence Analysis and System Optimization},'' \emph{{IEEE} Trans. Mobile Comput.}, 2025.

\bibitem{lu2024not}
X.~Lu, Q.~Liu, Y.~Xu, A.~Zhou, S.~Huang, B.~Zhang, J.~Yan, and H.~Li, ``{Not All Experts are Equal: Efficient Expert Pruning and Skipping for Mixture-of-Experts Large Language Models},'' in \emph{Proc. ACL (Volume 1: Long Papers)}, 2024, pp. 6159--6172.

\bibitem{fedus2022switch}
W.~Fedus, B.~Zoph, and N.~Shazeer, ``{Switch Transformers: Scaling to Trillion Parameter Models with Simple and Efficient Sparsity},'' \emph{Journal of Machine Learning Research}, vol.~23, no. 120, pp. 1--39, Apr. 2022.

\bibitem{muzio2024seer}
A.~Muzio, A.~Sun, and C.~He, ``{SEER-MoE: Sparse Expert Efficiency Through Regularization for Mixture-of-Experts},'' \emph{arXiv preprint arXiv:2404.05089}, Apr. 2024.

\bibitem{du2024sida}
Z.~Du, S.~Li, Y.~Wu, X.~Jiang, J.~Sun, Q.~Zheng, Y.~Wu, A.~Li, H.~Li, and Y.~Chen, ``{SIDA-MOE: Sparsity-Inspired Data-Aware Serving for Efficient and Scalable Large Mixture-of-Experts Models},'' \emph{Proc. MLSys}, vol.~6, pp. 224--238, Apr. 2024.

\bibitem{guo2021pfl}
B.~Guo, Y.~Mei, D.~Xiao, and W.~Wu, ``{PFL-MoE: Personalized Federated Learning Based on Mixture of Experts},'' in \emph{Proc. APWeb-WAIM}.\hskip 1em plus 0.5em minus 0.4em\relax Springer, Aug. 2021, pp. 480--486.

\bibitem{lewis2021base}
M.~Lewis, S.~Bhosale, T.~Dettmers, N.~Goyal, and L.~Zettlemoyer, ``{Base Layers: Simplifying Training of Large, Sparse Models},'' in \emph{Proc. ICML}.\hskip 1em plus 0.5em minus 0.4em\relax PMLR, Jul. 2021, pp. 6265--6274.

\bibitem{reisser2021federated}
M.~Reisser, C.~Louizos, E.~Gavves, and M.~Welling, ``{Federated Mixture of Experts},'' \emph{arXiv preprint arXiv:2107.06724}, Jul. 2021.

\bibitem{dun2023fedjets}
C.~Dun, M.~H. Garcia, G.~Zheng, A.~H. Awadallah, R.~Sim, A.~Kyrillidis, and D.~Dimitriadis, ``{Fedjets: Efficient Just-in-time Personalization with Federated Mixture of Experts},'' \emph{arXiv preprint arXiv:2306.08586}, Jun. 2023.

\bibitem{zhang2015character}
X.~Zhang, J.~Zhao, and Y.~LeCun, ``{Character-Level Convolutional Networks for Text Classification},'' \emph{Proc. NeurIPS}, vol.~28, Dec. 2015.

\bibitem{jhang2021challenges}
C.-J. Jhang, C.-X. Xue, J.-M. Hung, F.-C. Chang, and M.-F. Chang, ``{Challenges and Trends of SRAM-based Computing-in-Memory for AI Edge Devices},'' \emph{{IEEE} Trans. Circuits Syst. I, Reg. Pap.}, vol.~68, no.~5, pp. 1773--1786, May 2021.

\bibitem{zhang2023resource}
X.~Zhang and S.~Debroy, ``{Resource Management in Mobile Edge Computing: A Comprehensive Survey},'' \emph{ACM Comput. Surv.}, vol.~55, no. 13s, pp. 1--37, Jul. 2023.

\bibitem{zhu2024esfl}
G.~Zhu, Y.~Deng, X.~Chen, H.~Zhang, Y.~Fang, and T.~F. Wong, ``Esfl: Efficient split federated learning over resource-constrained heterogeneous wireless devices,'' \emph{IEEE Internet of Things Journal}, vol.~11, no.~16, pp. 27\,153--27\,166, 2024.

\bibitem{zhang2024towards}
J.~Zhang, S.~Vahidian, M.~Kuo, C.~Li, R.~Zhang, T.~Yu, G.~Wang, and Y.~Chen, ``{Towards Building the Federatedgpt: Federated Instruction Tuning},'' in \emph{Proc. ICASSP}.\hskip 1em plus 0.5em minus 0.4em\relax IEEE, Apr. 2024, pp. 6915--6919.

\bibitem{tishby2000information}
N.~Tishby, F.~C. Pereira, and W.~Bialek, ``{The Information Bottleneck Method},'' \emph{arXiv preprint physics/0004057}, Apr. 2000.

\bibitem{alemi2016deep}
A.~A. Alemi, I.~Fischer, J.~V. Dillon, and K.~Murphy, ``{Deep Variational Information Bottleneck},'' \emph{arXiv preprint arXiv:1612.00410}, Oct. 2019.

\bibitem{bisk2020piqa}
Y.~Bisk, R.~Zellers, J.~Gao, Y.~Choi \emph{et~al.}, ``{Piqa: Reasoning About Physical Commonsense in Natural Language},'' in \emph{Proc. AAAI}, vol.~34, no.~05, Feb. 2020, pp. 7432--7439.

\bibitem{zellers2019hellaswag}
R.~Zellers, A.~Holtzman, Y.~Bisk, A.~Farhadi, and Y.~Choi, ``{HellaSwag: Can a Machine Really Finish Your Sentence?}'' in \emph{Proc. ACL}, 2019, pp. 4791--4800.

\bibitem{hendrycks2020measuring}
D.~Hendrycks, C.~Burns, S.~Basart, A.~Zou, M.~Mazeika, D.~Song, and J.~Steinhardt, ``{Measuring Massive Multitask Language Understanding},'' \emph{arXiv preprint arXiv:2009.03300}, Jan. 2021.

\bibitem{mei2024fedmoe}
H.~Mei, D.~Cai, A.~Zhou, S.~Wang, and M.~Xu, ``Fedmoe: Personalized federated learning via heterogeneous mixture of experts,'' \emph{arXiv preprint arXiv:2408.11304}, 2024.

\bibitem{men2024shortgpt}
X.~Men, M.~Xu, Q.~Zhang, B.~Wang, H.~Lin, Y.~Lu, X.~Han, and W.~Chen, ``{Shortgpt: Layers in Large Language Models Are More Redundant Than You Expect},'' \emph{arXiv preprint arXiv:2403.03853}, Oct. 2024.

\bibitem{jiang2024mixtral}
A.~Q. Jiang, A.~Sablayrolles, A.~Roux, A.~Mensch, B.~Savary, C.~Bamford, D.~S. Chaplot, D.~d.~l. Casas, E.~B. Hanna, F.~Bressand \emph{et~al.}, ``{Mixtral of Experts},'' \emph{arXiv preprint arXiv:2401.04088}, Jan. 2024.

\bibitem{feng2025pm}
Y.~Feng, Y.-a. Geng, Y.~Zhu, Z.~Han, X.~Yu, K.~Xue, H.~Luo, M.~Sun, G.~Zhang, and M.~Song, ``{PM-MOE: Mixture of Experts on Private Model Parameters for Personalized Federated Learning},'' in \emph{Proc. WWW}, Apr. 2025, pp. 134--146.

\bibitem{jin2025moe}
L.~Jin, Y.~Zhang, Y.~Li, S.~Wang, H.~H. Yang, J.~Wu, and M.~Zhang, ``{MoE$^{2}$: Optimizing Collaborative Inference for Edge Large Language Models},'' \emph{arXiv preprint arXiv:2501.09410}, Jan. 2025.

\bibitem{he2024towards}
S.~He, D.~Dong, L.~Ding, and A.~Li, ``{Towards Efficient Mixture of Experts: A Holistic Study of Compression Techniques},'' \emph{arXiv preprint arXiv:2406.02500}, Mar. 2025.

\bibitem{zhan2024fedmoe}
Z.~Zhan, W.~Zhao, Y.~Li, W.~Liu, X.~Zhang, C.~W. Tan, C.~Wu, D.~Guo, and X.~Chen, ``{FedMoE-DA: Federated Mixture of Experts via Domain Aware Fine-Grained Aggregation},'' in \emph{2024 20th International Conference on Mobility, Sensing and Networking (MSN)}.\hskip 1em plus 0.5em minus 0.4em\relax IEEE, 2024, pp. 122--129.

\bibitem{hu2025fft}
G.~Hu, Y.~Teng, P.~Wu, and N.~Wang, ``Fft-moe: Efficient federated fine-tuning for foundation models via large-scale sparse moe under heterogeneous edge,'' \emph{arXiv preprint arXiv:2508.18663}, 2025.

\bibitem{yi2024pfedmoe}
L.~Yi, H.~Yu, C.~Ren, H.~Zhang, G.~Wang, X.~Liu, and X.~Li, ``pfedmoe: Data-level personalization with mixture of experts for model-heterogeneous personalized federated learning,'' \emph{arXiv preprint arXiv:2402.01350}, 2024.

\end{thebibliography}

\end{document}